\documentclass[fleqn,10pt]{SelfArx} 

\usepackage[english]{babel} 

\usepackage{lipsum} 

\definecolor{color1}{RGB}{0,0,90} 
\definecolor{color2}{RGB}{0,20,20} 

\usepackage{hyperref} 

\hypersetup{
	hidelinks,
	colorlinks,
	breaklinks=true,
	urlcolor=color2,
	citecolor=color1,
	linkcolor=color1,
	bookmarksopen=false,
	pdftitle={Title},
	pdfauthor={Author},
}

\JournalInfo{Arxiv Preprint} 
\Archive{Nov 18th, 2024} 

\PaperTitle{Signaling and Social Learning in Swarms of Robots} 

\Authors{Leo Cazenille\textsuperscript{1}, Maxime Toquebiau\textsuperscript{2,3}, Nicolas Lobato-Dauzier\textsuperscript{4}, Alessia Loi\textsuperscript{3}, Loona Macabre\textsuperscript{3}, Nathanaël Aubert-Kato\textsuperscript{5}, Anthony Genot\textsuperscript{6}, Nicolas Bredeche\textsuperscript{3}*} 
\affiliation{\textsuperscript{1}\textit{Université Paris Cité, CNRS, LIED UMR 8236, F-75006 Paris, France}} 
\affiliation{\textsuperscript{2}\textit{ECE Paris, France}} 
\affiliation{\textsuperscript{3}\textit{Sorbonne Université, CNRS, ISIR, F-75005 Paris, France}} 
\affiliation{\textsuperscript{4}\textit{Sorbonne Université, CNRS, IBPS, Laboratoire Jean Perrin, F-75005 Paris, France}} 
\affiliation{\textsuperscript{5}\textit{Department of Information Sciences, Ochanomizu University, Tokyo, Japan}} 
\affiliation{\textsuperscript{6}\textit{LIMMS (IRL2820)/CNRS-IIS, University of Tokyo, Tokyo, Japan}} 
\affiliation{*\textbf{Corresponding author}: nicolas.bredeche@sorbonne-universite.fr} 

\newcommand{\keywordname}{Keywords} 

\newcommand{\red}[1]{{\color{red}#1}}
\renewcommand{\red}[1]{#1}

\Abstract{This paper investigates the role of communication in \red{improving coordination within robot swarms, focusing on a paradigm where learning and execution occur simultaneously in a decentralized manner}. We highlight the role communication can play in addressing the credit assignment problem (individual contribution to the overall performance), and how it can be influenced by it. We propose a taxonomy of existing and future works on communication, focusing on information selection and physical abstraction as principal axes for classification: from low-level lossless compression with raw signal extraction and processing to high-level lossy compression with structured communication models. The paper reviews current research from evolutionary robotics, multi-agent \red{(deep)} reinforcement learning, language models, and biophysics models to outline the challenges and opportunities of communication in a collective of robots that continuously learn from one another through local message exchanges, illustrating a form of social learning.}

\begin{document}

\maketitle 

\tableofcontents 

\thispagestyle{empty} 



\section{Introduction}

A general and well-accepted definition of swarm robotics highlights the deployment of a possibly large collective of robots each with limited computation and communication capabilities working together \red{as a result of multiple local interactions} to achieve a common cause~\cite{beni1993swarm,dudek1993taxonomy,Brambilla2013,Hamman2018,dorigo2020reflections,floreano2021individual}. It is important to note that “limited” does not mean “simple”: a hypothetical collective of idealized self-aware language-capable robots could still be considered a swarm if decentralized coordination is required due to the inherent delay in communication, even if the environment is static. 
The limited capabilities of each robot are to be understood as a relative property that puts into relation two conceptual levels: (1) at the individual level, the capabilities of one individual component of the swarm, which encompass both its physical (sensors and actuators) and algorithmic (memory and computing power)  capabilities and (2) at the global level, the swarm complexity in terms of its size and spatial configuration, which define the possibilities of interactions between its components. While the hardware capabilities of the robots limit the goals that can be achieved, the limitation in software capabilities is the key factor. Whenever memory or computation is lacking at the individual level, collective action requires decentralized coordination\red{, as each robot can only sense and act in its immediate surroundings}. In addition, it is important to consider the time component of computational complexity, which depends on either or both a time-constrained task and an inherently dynamic environment. This implies that the swarm's response time should be short enough for its actions to be relevant.
 
\begin{figure*}[t]
\centering\includegraphics[width=0.90\textwidth]{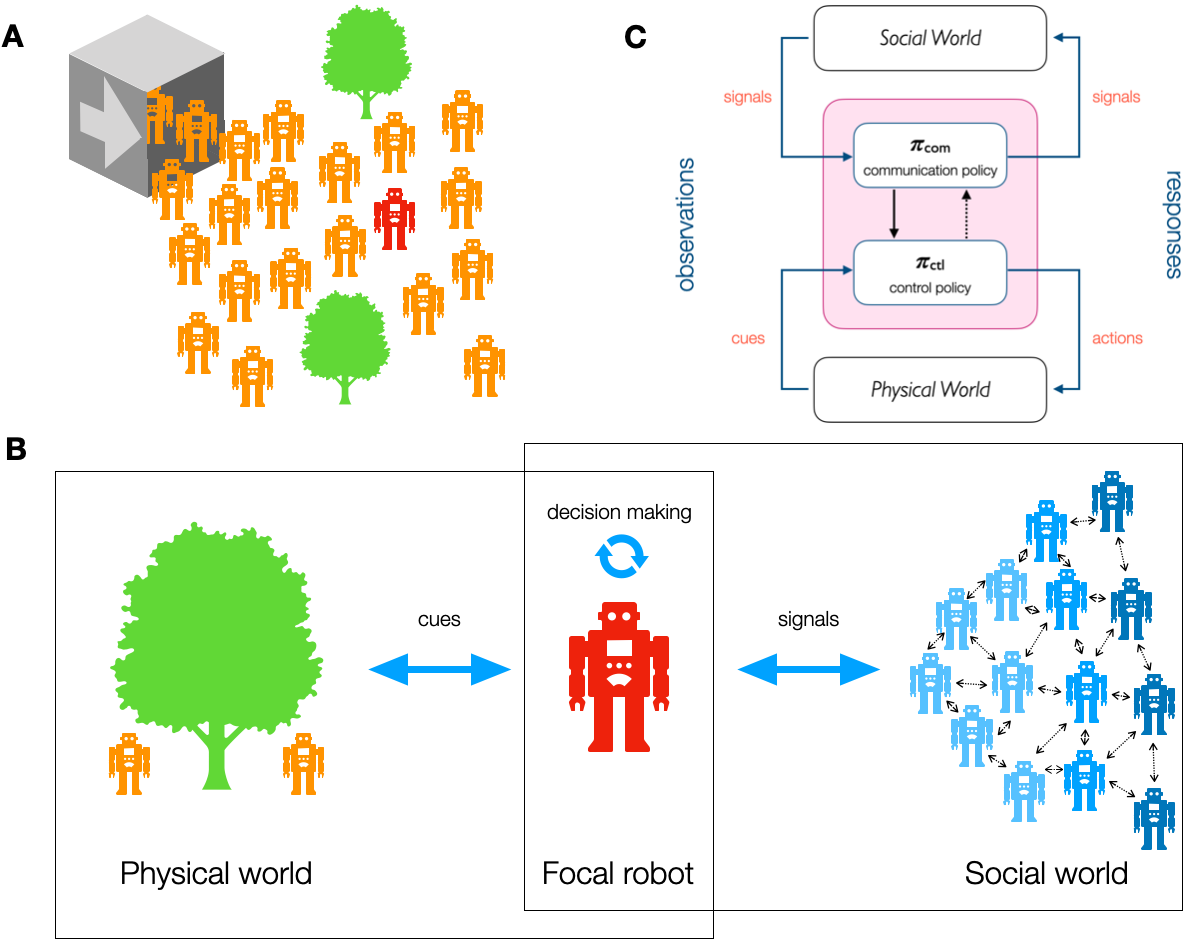}
\caption{(A) A swarm of robots is deployed in an unknown environment. Robots must learn together to solve a task. Robots interact locally with nearby robots and physical elements. (B) The decision-making process of a focal robot is based on cues from the physical world and signals from the social world. (C) Diagram of the communication and control policies for a robot, distinguishing between signals for local interactions and cues from the broader environment. The pink box denotes the policy of the robot which gets information from observations (i.e., cues and signals) and produces actions (i.e., effectors and communication channels). There are two sub-policies for each process, though in practice a single general policy may be used (e.g., a single artificial neural network), or multiple policies, either \textit{ad hoc} or subject to learning.}
\label{fig_general}
\end{figure*}

In this paper, we do not impose limitations over the actual capabilities of the robots or on the swarm structure (e.g., a heterogeneous swarm of unconventional robots is possible) and allow for different interpretations of what cooperation means (e.g., from just avoiding each other to displaying complex coordinated strategies). This is covered by this slightly different and more accurate definition: \textbf{\red{swarm robotics involves deploying robotic agents that coordinate in a decentralized manner to achieve a common goal, with each robot limited to sensing and acting within its immediate environment\footnote{We consider swarms of self-interested robotic agents as off-topic from the present study}}}. That definition opens new venues for thinking about the future of the field, including bridges toward other fields with similar concerns, as we will see later.

The design of efficient individual policies within a swarm of robots usually relies either on carefully crafting (possibly bio-inspired) behavioral rules or on using learning and/or evolutionary optimization algorithms. The robots’ policies, which are generally similar across a given swarm, do not change after deployment. While this approach is enough in many cases, it is a limitation whenever the target environment is unknown before deployment or changes over time. This is why an important effort, originally stemming from evolutionary robotics, has been made since the advent of the 21st century to develop decentralized online evolutionary learning algorithms. This family of algorithms aims at enabling a robot swarm to adapt continuously while already deployed in the real world, as illustrated in Figure~\ref{fig_general}-A, and has been referred to as either embodied evolutionary (EE) robotics~\cite{watson2002embodied} or social learning for swarm robotics (SLSR)~\cite{heinerman2015evolution,bredeche2022rspt}. They have achieved remarkable success in terms of the number of implementations on real robots when compared to other fields working with learning multi-robot systems (see~\cite{bredeche2018ee} for a review). 

In this paper, we posit that the class of problems addressed when using such social learning or embodied evolution algorithms is covered by the umbrella term of \textbf{decentralized learning and execution (DLE)}, which designates a paradigm that will be more familiar to the wider reinforcement learning community~\cite{zhang2018fully,lyu2021contrasting}. This contrasts with the widely used ``design then deploy'' paradigm, which includes (1) prior hand design, (2) offline evolutionary robotics, and (3) multi-agent reinforcement learning under the centralized training and decentralized execution paradigm.

We focus here on \textbf{how communication within a swarm of robots can be used to improve coordination under the DLE paradigm}. Communication can obviously be used by each robot in the swarm to enable information sharing and/or synchronized behavioral response~\cite{Brambilla2013,Hamman2018}. However, communication can also play a role in the very nature of the learning process as all computation regarding learning under the DLE paradigm is performed on the field, without any human or central computer involved. \red{A straight-forward} form of communication in this context is exemplified by the crude control parameter sharing at work in embodied evolutionary algorithms~\cite{ficici1999embodied,watson2002embodied,bredeche2010ppsn,bredeche2022rspt}, where (either all or a sub-part of) the neural weights of Artificial Neural Networks are sent from one robot to its neighbors, possibly attached with a self-assessment of its performance from local observations (details in~\cite{bredeche2018ee}). 

In Section~\ref{sec-dynamics}, we start by exploring how working in the DLE paradigm raises unique challenges, whether communication among robots is enabled or not. We expose how using DLE can lead to counter-intuitive consequences due to learning in a decentralized fashion, in particular regarding unwanted and counter-productive competition among robots. In Section~\ref{sec-taxonomy}, we propose a taxonomy to characterize existing and future works on communication. Section~\ref{sec-review} offers a review of existing works that draw from several very different domains such as biophysics, evolutionary robotics, language evolution, multi-agent deep reinforcement learning, and language models~--~to provide an overview of current and future directions. We propose a classification of communication means along the axes of information \red{selection} and physical abstraction, ranging from raw information directly available in the environment in Sec.~\ref{sec-review-Lowdimred} (e.g., transfer of heat or matter, either as raw information or as mathematical abstractions) up to high-level language-based communication in Sec.~\ref{sec-review-Highdimred} (e.g., emergent or human-like syntax and grammar). 
Finally, the last Section summarizes important ideas explored earlier and provides comments and considerations for the future.


\section{Dynamics of Decentralized Learning and Execution}
\label{sec-dynamics}
As stated in the Introduction, natural evolution and social learning are good examples of processes working under the DLE paradigm. Individuals compete with one another to gain a selective advantage. Combined with random variations and inheritable traits, the traits of successful individuals will become more frequent over time. Of course, there is a stark contrast between natural systems and swarm robotics systems: we engineer the robot swarm to address a particular problem defined before deployment which may require coordination to be addressed (foraging, exploration, patrolling, transporting, construction, or monitoring to give a few examples~\cite{schranz2020swarmapplis}). While the desired outcome may be relatively easy to define, the challenge is to endow each robot with the capability to assess how much it contributes to solving the task, i.e., self-assessing the robot’s contribution to the global welfare of the collective, which is itself determined by how efficiently the task is solved. 

In a collective, devising the contribution of each individual is referred to as the credit assignment problem, which is well known in the multi-agent and cooperative game theory communities~\cite{Nisan2007agt,oroojlooy2023review}. If a complete alignment of the individual’s interest with the global welfare of the collective is possible, the best actions from the robot's viewpoint will also be the best for the collective. In a setup where individual policies are learned, this corresponds to converging towards a Nash Equilibrium that is also a social optimum, meaning none of the robots has the incentive to deviate from its current behavioral strategy as it is already the best the robot can do reward-wise (\red{see~\cite{wolpert1999introduction,stone2010ad,marden2018game} for theoretical considerations in distributed robotic systems, and~\cite{ecoffet2021nothing} for a practical example with evolutionary learning in a swarm of robots where there is a mismatch between evolutionary stable strategies and social optimal strategies}).

A direct way to make individual interests coincide with that of the team would be to provide each individual with a measure of their contribution to the global performance. However, \textbf{estimating the marginal contribution of each robot to the performance of the collective is intractable in the general case}. Even in an idealistic setting, when a scenario can be replayed an indefinite number of times and robots can be removed or added at will, \red{computation time for estimating the marginal contributions for each individual grows exponentially with the population size as all subsets of individuals must be considered~\cite{Shapley1953,shoham2008multiagent,wooldridge2009introduction}}. It is also interesting to note that the more classic reinforcement learning methods using centralized learning do not yield optimal results, as the robots’ marginal contributions are often partially or badly estimated even by the centralized critic used in multi-agent (deep) reinforcement learning~\cite{lyu2021contrasting}. One efficient simplifying hypothesis used in the field of evolutionary collective robotics is to consider a swarm of clones~\cite{Trianni2006,waibel2009genetic}, turning what originally looks like a collective decision-making problem into an optimization problem as a single control parameter set is used for the entire swarm and optimized in a centralized fashion. This method is however not applicable under the DLE paradigm as it requires a centralized coordinator for learning.

Approximation methods to estimate on the fly the marginal contribution of robots in a collective exist, of course, and trade tractability against a lack of optimality or assume simplifying hypotheses on the class of problems to be addressed (see in particular~\cite{Wolpert2000OptimalWL,kolpaczki2024approximating,wang2022cooperative}). A straightforward method is for the human supervisor to define \textit{a priori} an explicit evaluation function embedded in each robot whose goal is to evaluate locally the performance of said robot. This is the case with most works in embodied evolution and social learning in swarm robotics, where each robot computes an estimate of its performance based solely on directly available information and self-assessment~\cite{bredeche2018ee}. This is also the case in cooperative multi-agent learning whenever each agent is an independent learner, i.e., considering others as part of a nonstationary environment~\cite{gronauer2022multi}. In both cases, the global performance will depend on the ability of the human engineer to design a function that provides a reliable estimate of the performance of a robot, aligning local motivation with the desired global outcome. Obviously, this can quickly become challenging as the task and/or the environment grow in complexity — e.g., foraging in a field without obstacles can be very different from foraging in a complex environment where the division of labor offers a significant advantage.

Unfortunately, \textbf{the slightest misalignment between the individual’s interests and that of the collective can lead to a suboptimal group-wise performance}. In that case, the whole swarm will eventually converge towards a Nash Equilibrium that does not guarantee social optimality. This is explained by the nature of the evolutionary dynamics at work behind social learning in a swarm: elements that play a part in the robots’ behavioral strategies are competing among themselves to invade the population of robots. If the metric used to compare those elements is aligned (resp. not aligned) with the global task, then competition will end up with individual strategies that are optimal (resp. sub-optimal) w.r.t. the task. This can be explained by using the famous selfish gene metaphor popularized by Richard Dawkins~\cite{dawkins2016selfish}: robots are merely vehicles for competing units (e.g., genes or group of genes, neural network parameters, symbols from an emerging language, elements of an artificial culture, etc.) facing selective pressure.

Such evolutionary dynamics can then have a direct impact on the long-term behavioral strategies of neighboring robots, with sometimes surprising outcomes such as mutualistic cooperation (i.e., cooperation that benefits each involved party) and altruistic behavior (i.e., cooperation that involves a net loss at the individual level, but which indirectly benefits the survival of related individuals)~\cite{west2007social}. In particular, \textbf{each individual’s strategy is shaped by its inclusive fitness that captures both its ability to survive \textit{and} its ability to help related individuals} (a relation that is generally, but not always, defined at the genotypic level)~\cite{hamilton1964}. Kin selection, the process by which an individual favors their relatives, is also relevant for the development of cultural adaptation~\cite{richerson2008not} and language~\cite{smith2010communication}. This has been shown previously to also be the case with social learning algorithms for swarm robotics~\cite{montanier2011surviving}: robots can lose part of their survival chances to help robots with whom they share information.

\begin{figure}[t]
\centering\includegraphics[width=8.5cm]{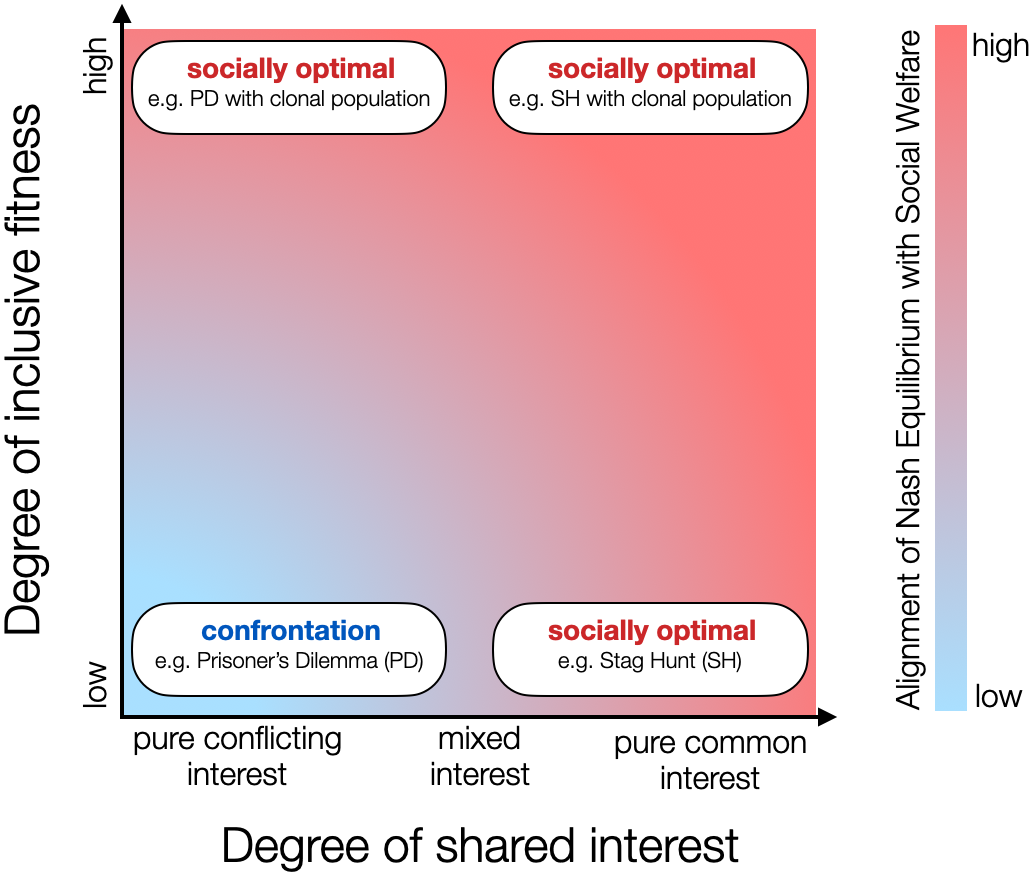}
\caption{Alignment of Nash Equilibrium with Social Welfare with respect to the degree of inclusive fitness and the degree of shared interest among robots. The X-axis shows how aligned the individual’s interest (e.g., its local fitness function) is with that of the group, which is uniquely defined by its ability to optimally solve the task. The Y-axis shows the level of inclusive fitness experienced by each individual in the population (e.g., due to kin recognition, environmental viscosity, etc.). 
The four text boxes on the graph provide examples using the well-known theoretical games of Prisoner’s Dilemma (PD, a competitive game where players should defect) and Stag Hunt (SH, a coordination game where players should cooperate), and two extremes regarding how inclusive is an individual’s fitness in a population (unrelated individuals working for their own sake vs. a population of clones working for the collective).}
\label{fig_alignment}
\end{figure}

Figure~\ref{fig_alignment} puts together the two concepts just discussed: (1) the stronger the alignment between the individual’s interest and the group’s welfare, the better the performance w.r.t. to the user-defined objective (x-axis) and (2) inclusive fitness, which shows the degree to which an individual's interest is aligned with that of its relatives (y-axis). 
\red{In this Figure, we oppose two extreme configurations, one in which individuals in the swarm are in confrontation with conflicting interests, and another one in which individuals' interests are aligned and individuals cooperate to maximizing the social welfare, whether this incurs an individual cost or not.}
Obviously, the level of cooperation for solving the user-defined task will be maximal if alignment is complete and may decrease otherwise depending on the task at hand. Much less obvious is the influence of inclusive fitness\red{, where an individual cost may be paid for the benefit of the whole. The intuition can be given by looking at the example of  eusocial} colonies (e.g., ants and termites) where the fitness of one individual is vastly defined by that of its \red{superorganism}. In that case, individual actions that benefit the group will be performed, even if they are detrimental to the individual (see~\cite{waibel2011quantitative} for a study of the impact of inclusive fitness in evolutionary collective robotics). To some extent, a high level of inclusive fitness can compensate for a misalignment between the individual’s interest and that of its conspecifics. In this Figure, we formulate this relation as the degree to which the Nash Equilibrium of the evolving population will converge to the socially optimal outcome with respect to the user-defined task.

We now turn back our attention to communication in a swarm of robots and the implication of previous considerations on it. Communication can be used to endow each robot with the ability to locally estimate its contribution in an online fashion, as shown by recent works in the field of cooperative multi-agent reinforcement learning that proposed using communication between robots to locally aggregate the data available on the performance of the swarm as a whole~\cite{zhang2021decentralized,zimmer2021learning}. In that way, communication can be used to gather data on a macroscopic scale so that more information is available to each individual regarding the performance of the whole, and possibly to provide an individual’s ability to measure its contribution. Although this does not solve the credit assignment problem, communication can help to perform counterfactual reasoning to simulate hypothetical scenarios in the absence of the focal robot~\cite{foerster2018counterfactual}.

Unfortunately, communication also suffers from a possible misalignment between the Nash Equilibrium and socially optimal strategies, especially if it evolves (e.g., emergent signaling or language).
In case of misalignment, environmental contingencies and competitive pressure among individuals can lead to sub-optimal communication strategies, as evolving communication undergoes the same pressures as learning the action policy, resulting in robots developing sub-optimal communication efficiency to gain a competitive advantage against competitors~\cite{wischmann2012historical}.
In turn, evolving communication may benefit from robots with a higher degree of inclusive fitness and/or a shared interest between individuals~\cite{floreano2007evolutionary}.


\section{A Taxonomy of Signaling Methods in Swarm Robotics}
\label{sec-taxonomy}
First, let us start by narrowing the scope regarding the nature of communication we are interested in by distinguishing cues from signals. Cues provide information to the focal individual, extracted from the environment through direct observations  (e.g., the relative alignment of nearby conspecifics~\cite{reynolds1987flocks,vicsek1995novel}) or identification of body markers (e.g., a conspecific's phenotypic trait). They do not require an identified interlocutor and, if another individual is involved, they are not produced intentionally. Signals involve an emitter and at least one receiver. They are produced intentionally by the emitter through one or several available modalities (auditory, visual, olfactory, etc.), and can vary greatly in complexity, from the production of a chemical compound to human language. The interested reader can refer to~\cite{smith2003animal} for a comprehensive introduction to cues and signaling in nature.

Figure~\ref{fig_general}-B and~-C provides an illustration from a robot swarm perspective. Each robot may experience both cues, observed in the physical world, and signals, originating from other robots and received through dedicated channels such as short-range proximity communication devices (e.g., infrared, visible light, radio, etc.). We explicitly limit our scope to the moment when information from the signal is readily available to the robot, leaving any pre-processing transparent (signals can be initially extracted from another modality such as speech and sign language, as is the case in robot-human communication~\cite{Berghe2018socroblg}).

Communication strategies in swarm robotics cover both stigmergic communication and direct communication. Stigmergic communication works by leaving a trace in the environment~\cite{bonabeau2000inspiration,detrain2008collective}, such as a virtual pheromone trail for other robots to consider~\cite{bonabeau1999swarm,campo2010artificial}). Direct communication involves explicit exchanges of information among robots, either through pre-defined or emergent signaling strategies. In particular, emergent communication strategies evolve naturally from the interactions and the optimization processes at work within the swarm, enabling robots to converge towards efficient adaptive behaviors without centralized control.

In addition to whether signaling strategies are learned or pre-defined, the nature of the signals can vary greatly taking, for instance, discrete and continuous forms. Low-level communication methods often mimic natural processes like diffusion, reaction, and advection, enabling robots to share information about their local environment. High-level methods involve more abstract forms of communication, such as emergent or structured language models, allowing for sophisticated interactions and decision-making. 

\begin{figure*}[t]
\centering
\begin{tabular}{c}
\includegraphics[width=14.5cm]{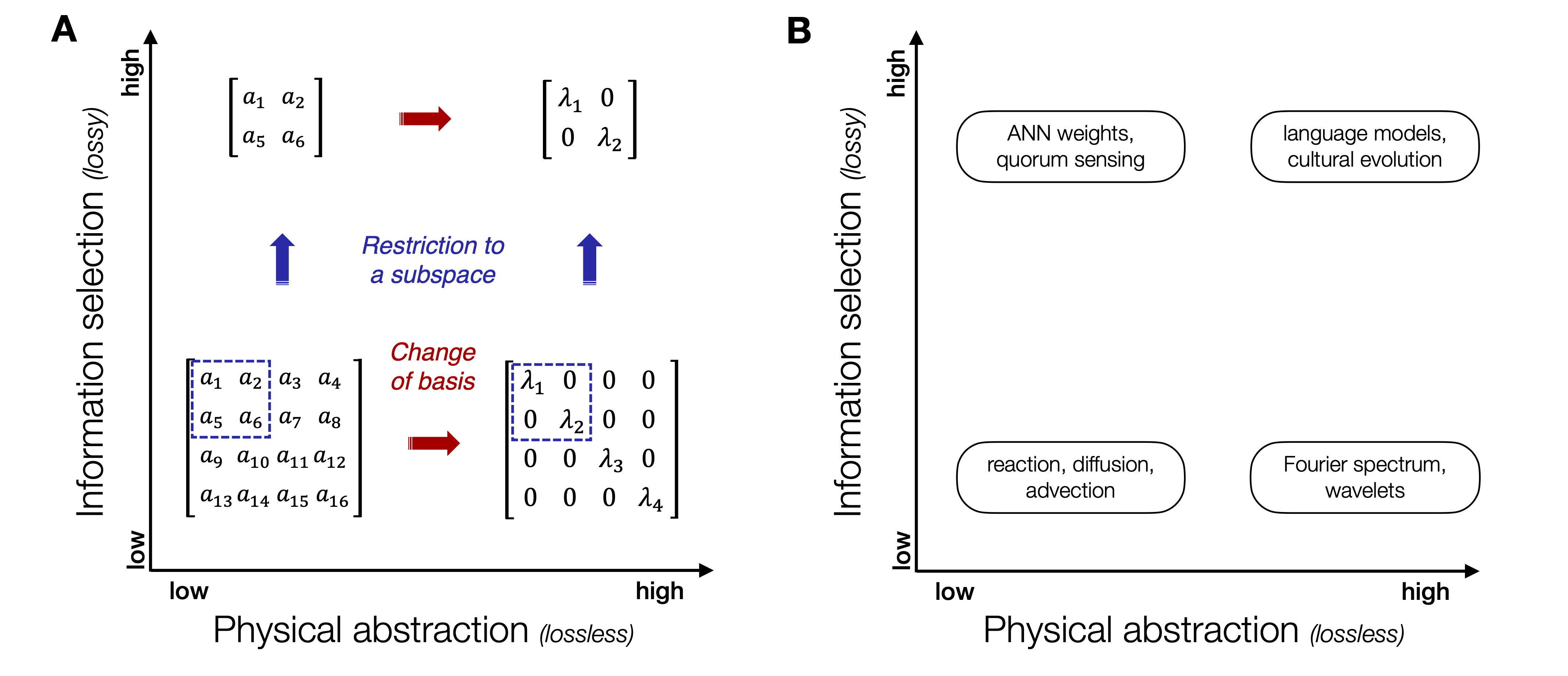} \\
\end{tabular}
\caption{Signaling methods can be projected in a two-dimensional plane using information \red{selection} and physical abstraction as main components. \textbf{Left}: An algebraic analogy for information \red{selection} and physical abstraction in communication processes. Changing the degree of information \red{selection} can be done through operations like restriction to a subspace (projection with or without loss). Changing the level of physical abstraction can be done via a change of basis, such as transforming a complex matrix into a simpler diagonal form, illustrating how information can be simplified and structured. \textbf{Right}: Different approaches to communication in robotics, mapped by information \red{selection} and physical abstraction. Low-level methods include biophysics-inspired processes, while high-level methods involve language models and emergent languages.}
\label{fig_communication}
\end{figure*}

\red{Signaling also necessarily incurs some form of restriction over the nature and the amount of information that will be shared, driven by the necessity to transfer relevant information only. This process may be lossless (e.g.,~suppressing redundant information, compressing information without loss, or changing the way information is represented) or lossy (e.g.,~ignoring irrelevant information, compression with loss). In practice, as the complexity of the environment increases, so does the need for sharing only what is relevant for the task at hand (e.g., selection attention in humans~\cite{carrasco2011visual}, or methods used to avoid the curse of dimensionality in machine learning~\cite{bellman1966dynamic}).

We propose two axes for classification using the \textbf{degree of information selection} and the \textbf{degree of physical abstraction}. On the one hand, information selection aims at reducing the quantity of information shared by loosing information that is not deemed relevant. On the other hand, physical abstraction aims at changing the way information is represented without loss of information in order to reveal what is already present.}
This is illustrated in Figure~\ref{fig_communication}. The left part of the figure provides an analogy with algebra to provide an intuition using a mathematical metaphor. The right part maps well-known approaches used in swarm and collective robotics, which will be explored further in the later Sections.

In the region considering a low degree of both information \red{selection} and physical abstraction, communication processes are closely tied to raw physical phenomena, such as reaction, diffusion, and advection. These methods mimic natural processes to transfer information, focusing on detailed, low-level interactions. Increasing the level of physical abstraction (x-axis) enables the extraction of hidden but highly relevant information from raw information such as using spectrum analysis or Fourier transforms, e.g., to capture geometric information of collective spatial configurations~\cite{cazenille2024hearing}. Similarly, a second axis (y-axis) explores how increasing the degree of information \red{selection} can extract relevant information in different forms. For example, sharing parameters of an artificial neural network controller that maps sensory inputs to motor outputs, as is common in adaptive swarm robotics~\cite{bredeche2018ee}, can be seen as a highly compressed (and biased) instance of a reaction-diffusion process.
Finally, the extreme of both axes points towards signaling strategies with high degrees of information \red{selection} and physical abstraction, where we can find, e.g., the use of a communication apparatus that is based on a large language model, enabling human-level structured perception and signaling.


\section{\red{Current Trends in} Signaling \red{for} Swarm Robotics}

\label{sec-review}

In this Section, we provide a review of relevant methods from swarm robotics as well as other domains, to reveal what the future states of signaling could be, considering both \textit{ad hoc} and emerging signaling methods. The Section follows the structure provided earlier: we first describe signaling methods with a low degree of information \red{selection} in Sub-section~\ref{sec-review-Lowdimred}, then move up to those with a high degree of information \red{selection} in Sub-section~\ref{sec-review-Highdimred}. In each Sub-section, strategies with different degrees of physical abstraction are described, also drawing from domains beyond that of swarm robotics. We make significant room for signaling methods used in multi-agent reinforcement learning as well as in the currently popular domain of large language models (LLMs). As mentioned in the Introduction, we stress that while existing swarm robotics hardware is still technically limited, this state of affairs may change in the near future. As a consequence, we expect that collective systems that can be identified under the umbrella of robot swarms will feature embedded computation capabilities powerful enough to run, and possibly train in real-time, LLMs (e.g.,~LLMs can already run on limited hardware~\cite{gerganov_llama_cpp}).

\subsection{Low Degree of Information \red{Selection}}
\label{sec-review-Lowdimred}

\textit{\textbf{Summary:} In this section, we investigate communication in multi-agent systems as information exchanges with minimal simplification of the baseline observable data from local agents. This involves two types of signaling schemes.
Scheme (1) involves signals that reflect local observations directly. Biological examples include social insect communication and autoinducers exchanges in bacteria. In swarm robotics, these principles are applied through algorithms mimicking biological behaviors through local interaction and communication rules. 
Scheme (2) includes signals with a high level of physical abstraction and structure, such as the use of Fourier transforms and wavelets to analyze and share periodic patterns and multi-scale features in data. In swarm robotics, agents might \red{share with immediate neighbors their} computed gradients or neural network weights, or perform Fourier transforms or eigenspectrum analysis to understand and communicate the underlying structure of complex data. }

Multi-agent communication with low information \red{selection} (lower part of Figure~\ref{fig_communication}) involves the direct and explicit exchange of observable information from local agents~\cite{ji2023signal}. The signals are transmitted in a form that retains most of the original observations, without any extensive selection mechanism removing parts of the baseline observation data. In Figure~\ref{fig_general}C, communication with low information \red{selection} involves minimal loss of information between the cues from the environment and their packaging into signals sent to other agents.
This approach is relevant either 1) in cases where the observations already have low dimensionality, 2) in cases where most of the observations contribute to the collective dynamics of the group, or 3) in cases where knowing in advance which parts of the observations are useful to communicate is difficult to achieve.

\textbf{(1) Low physical abstraction:}
In the case with both low information \red{selection} and low physical abstraction (lower left quadrant in Figure~\ref{fig_communication}), signals represent direct and tangible information about the environment, with minimal transformations from the observations of local agents.
This type of signaling is exemplified by the following biological systems:
autoinducers exchanges among bacteria~\cite{taga2003chemical}, auditory and tactile signals in Drosophila~\cite{larue2015acoustic}, chemical alarms released from certain fish species to alert conspecifics of the presence of a predator~\cite{bairos2019novel}, the bioluminescence mechanisms of fireflies for mate attraction~\cite{marques2009firefly}, electric signals in certain fish~\cite{hopkins1974electric}, or birds songs to attract mates~\cite{eriksson1986male} or to signal aggressive intent~\cite{searcy2006bird}.

Swarm robotics algorithms deployed on small robots or with self-organization capabilities also fit in this quadrant, because they rely on simple ad hoc signaling rules based directly on local states and observations, without significant loss of information or transformations. 
For instance, in~\cite{floreano2007evolutionary} a signaling behavior is optimized so that robots emit specific signals when they are close to an object or zone of interest. In ~\cite{rodrigues2015overcoming} robots share all their local sensory information with their neighbors during a predator-prey task.
The relative position of each robot or site of interest is locally broadcasted in~\cite{talamali2021less,mcguire2019minimal}.
In~\cite{hafnaoui2019timing}, robots can probabilistically broadcast information from one to another to assess the dynamics of information propagation.

Multi-agent systems inspired \red{by} physical dynamics can \red{also} be classified in this category: e.g., reaction-diffusion~\cite{crank1979mathematics}, \red{chemical oscillations~\cite{shanks2001modeling} and morphogenesis~\cite{turing1990chemical}} can \red{be seen} as multi-agent systems where agents are spatial discretization points and global dynamics emerge from local interactions \red{(communication without \red{information} loss)}.
Diffusion is a fundamental physical process where particles spread from areas of higher density to areas of lower density. In multi-agent systems, diffusion can serve as a means of communication. For example, chemical signaling in \red{cells} relies on \red{molecular} diffusion to \red{guide} movement, growth, and specialization.
Reaction-diffusion systems \red{involve the creation, transformation, or destruction of diffusive elements} through local interactions \red{to} create complex patterns. In multi-agent \red{systems}, reaction-diffusion can explain how agents interact with their environment and each other \red{via} chemical signals~\cite{arai1993information}.

\red{Moreover, making robots out of molecules allows the creation of massive swarms of millions of robots. In the last decades, researchers have used artificial DNA as computing and building blocks to develop molecular robotics~\cite{murata2022molecular}. Such robots can take the form of DNA origami that self-assemble into complex 3D nanostructures, able to connect to each other or change configuration depending on biochemical cues~\cite{nummelin2020robotic,douglas2012logic,torelli2014dna,kuzuya2014nanomechanical,amir2015folding,kaminka2017molecular,daljit2018switchable,li2018dna}. Simpler structures can also be programmed to move on tracks~\cite{wickham2012dna} and sort cargoes at the nanoscale~\cite{thubagere2017cargo}. Coating beads with DNA allows to create micro-robots with higher computing capabilities, with reaction-diffusion serving to form both controllers and signals~\cite{gines2017microscopic,aubert2017evolutionary,cazenille2019exploring}}. \red{Another emerging field is controllable active matter, where self-propelled agents process chemical signals locally, leading to self-organization}~\cite{ziepke2022multi,wang2024swarm,grauer2024optimizing,lavergne2019group,keya2018dna,akter2022cooperative,aubert2023collective}.
\red{A final example is the Turing model of morphogenesis, which explains how patterns like animal stripes and spots emerge from the interaction of diffusing chemicals, inspiring a swarm robotics implementation where local communication mimics a reaction-diffusion system to achieve shape formation~\cite{slavkov2018morphogenesis}.}

\textbf{(2) High physical abstraction:}
The lower right quadrant of Figure~\ref{fig_communication} represents communication methods involving abstract and structured information, often detached from direct physical processes, with \red{minimal} information loss from observations.

This includes methods \red{like} broadcasting gradients \red{where agents locally exchange mathematical abstractions} rather than direct physical signals. Gradients represent \red{the internal} state of each agent's model, rather than a direct physical quantity.
For instance, gradient propagation can \red{compute} a geodesic distance to a source robot \red{by incrementally communicating values through neighboring agents}~\cite{rubenstein2014programmable}.
\red{Gradient broadcasting can occur through} microscopic rules \red{derived from local observations}~\cite{gauci2017error,wang2020fast,rubenstein2014programmable}, or \red{via} multi-agent reinforcement learning where the gradients of the loss function are broadcasted from agents to agents~\cite{busoniu2008comprehensive}. \red{Having differentiation capabilities, i.e., access to the gradient of local states and/or messages, allows the training process to directly use this information (e.g., via gradient descent algorithms), accelerating convergence.}

\red{Eigenspectrum analysis~\cite{saad2011numerical} also fits this quadrant, examining eigenvalues and eigenvectors to reveal the underlying structure of data.} Eigenspectrum analysis is widely used in a variety of fields, ranging from signal processing and machine learning to network analysis.
This process can involve similar dynamics as those obtained in the lower left quadrant -- however, it will also use mathematical tools to change the representation of information without loss of information.
For instance, in~\cite{cazenille2024hearing} a swarm of Kilobot robots estimates, in a decentralized way, the eigenspectrum of the communication graph between robots. Such properties are then used to reach a global consensus on the shape of the swarm, achieving arena shape recognition. This process is achieved by relying on a physics-inspired communication scheme based on the diffusion of heat across the swarm and mathematical tools to \red{locally} extract the second eigenvalue $\lambda_2$ of the graph Laplacian, a direct fingerprint of the arena shape containing the swarm.

\red{Fourier transforms~\cite{oppenheim1999discrete} and wavelets are other examples of abstract tools. Fourier transforms convert signals between the time and frequency domains, enabling agents to analyze and share information about periodic patterns. Wavelets decompose data into different scales, allowing agents to communicate detailed features of a signal, from broad trends to fine details.} While Fourier transforms and wavelets are not yet used to process signals in swarm robotics settings, their capabilities to work with more abstract representations may allow a new class of communication schemes -- e.g., to perform distributed spectral analysis as a result of communication, as in~\cite{cazenille2024hearing}.

\subsection{High Degree of Information \red{Selection}}
\label{sec-review-Highdimred}
\textit{\textbf{Summary:} In this section, we explore decentralized communication as viewed from the prisms of the information bottleneck, language evolution, and multi-agent reinforcement learning in situated environments. Reinforcement learning approaches to emergent communication are examined, highlighting both benefits and challenges. We emphasize the \red{opportunities provided by LLMs for advancing communication in swarm robotics, noting their strengths in generating human-like language and reasoning, and} challenges such as biases, hallucinations, embodiment, and efficient deployment on \red{robots. Overall, we present a range of approaches, with different degrees of physical abstraction, that enable decentralized agents to learn communication.}}

\red{In realistic decentralized environments, all observed information is not relevant to transmit to partners. Thus, a higher degree of information selection (upper part of Figure~\ref{fig_communication}) is required to allow efficient transmission of the relevant information. }
This task can be decomposed into two complementary sub-tasks: (1) selecting relevant information and (2) transmitting this information.
The selection task involves extracting parts of the observed information that are relevant to other agents. The transmission task requires coding this information so other agents understand it while ensuring that bandwidth constraints are respected.
Both tasks are highly interconnected. The selected information requires adequate means of coding to be transmitted without (or with minimal) loss. The transmission means, in turn, influence the information selection by dictating what information can be transmitted efficiently~\cite{Ohmer2022:MutualInfluence}.
This is a form of information bottleneck, where agents need to generate a compressed mapping of their observations, that contains as much information as possible related to the task at hand~\cite{Tishby1999:IB}.
As communication comes necessarily at a cost, languages operate a trade-off between \red{meaning and compression}~\cite{Kirby2015_CompressionComm, Zaslavsky2018:ColorNamingEvo}\red{, maximizing expressiveness while minimizing communication costs.}


\red{Studying \textit{language games} shows how languages emerge from this information bottleneck, and from various ecological constraints. }
\red{In his seminal work, Luc Steels}~\cite{Steels1999:TalkingHeads1} \red{demonstrated that having} a dynamic population of \red{embodied} agents, whose reasoning is unknown \red{to one another}, motivates the emergence of a shared compositional language. 
In the iterated learning framework~\cite{Kirby2001:ILM,Smith2003:IteratedLearning}, the emphasis is put on \red{a transmission bottleneck that occurs when language is transmitted between successive generations of agents, driving languages to adopt simple and compositional structures.} 
\red{Following works have shown that emergent languages are also shaped by environmental~\cite{Vogt2005:EmergenceCompo, Perfors2014_WorldShapeLang} and physiological~\cite{Christiansen2008_BrainShapeLang} constraints. These experiments highlight the different requirements for languages to originate in populations of independent agents, and demonstrate the emergence of efficient naming and grammatical conventions~\cite{Wellens2008:FlexibleWordMeaning, Beuls2013:GrammaticalAgreement, Zaslavsky2018:ColorNamingEvo, Ekila2024:LinguisticConventions}.} 

However, these language games still heavily simplify the context of communication interactions, by making the agents, their observations, and their actions, solely defined by the communication game. Previous works have classified this kind of setting as \textit{non-situated}~\cite{Wagner2003:EC}, as opposed to \textit{situated} agents that have a localized existence and can physical\red{ly interact with} their surroundings. In situated environments, communication is one of many interfacing processes. It can be used for communicating not only about observations, but also about intents, or even about task-agnostic and abstract concepts. It may involve non-cooperative agents. It might not even be required at all times. In such realistic settings, choosing which information is relevant to communicate is a much more complex task that involves reasoning about the current state of the environment, the agent’s objective, and the current knowledge and reasoning of other agents. In that sense, learning to communicate is inherently a multi-agent problem of learning how to behave in a dynamic, partially observable environment. 

Recently, research in multi-agent reinforcement learning has tackled such situated environments, where performance depends on a combination of physical and communication behavior~\cite{Zhu2024_MACSurvey}. In this context, multi-agent systems learn, often with centralized training and decentralized execution, to generate messages that participate in maximizing future returns. \red{Here, messages are continuous vectors generated by neural networks inside the agents' system. This makes} communication a \textit{differentiable sub-step of the action selection process}, which can be learned fully end-to-end as a tool \red{for maximizing returns}~\cite{Sukhbaatar2016:CommNet, Foerster2016:DIAL, Peng2017:BiCNet, Wang2022:FCMNet}. Because the message generation is differentiable, gradients can flow between agents. Thus, messages are \red{explicitely} trained to help other agents maximize their rewards. This approach has been extended in various ways for more targeted information sharing~\cite{Hoshen2017:VAIN, Jiang2018:ATOC, Das2019:TarMAC} or to limit bandwidth usage~\cite{Singh2019:IC3Net, Zhang2019:VBC, Wang2020:IMAC, Han2023:MBC}. Similar approaches have been developed using discrete symbols for communication~\cite{Cao2018:Negotiation, Lazaridou2018:Emergence, Jaques2019:SocialInfluence, Kim2018:SchedNet,Rita2022:GenOverf}. In those, agents have to reach a consensus on the meaning of each symbol through trial and error. The compression constraint depends both on the size of the vocabulary and the size of the sequences. Previous works have shown that imposing constraints on both of these attributes induces emergent languages to develop common characteristics of natural languages such as compositionality~\cite{Mordatch2018:GroundedCompo, Rita2022:PopHetero} and abbreviation of frequent words~\cite{Chaabouni2019:AntiEfficient}.

However, learning emergent communication through this task-oriented reinforcement learning process has many important limitations. As already mentioned, it often requires a centralized \red{learning} algorithm to allow reinforcement learning tools to reliably converge to adequate solutions. As with all gradient-based learning methods, it acts as a black box that lacks practical ways of interpreting and measuring its efficiency~\cite{Lowe2019:Pitfalls, Lazaridou2020_DeepEmergentComm}. \red{More importantly, differentiable emergent communication gives no guarantee of learning to communicate about concepts from the environment. Rather, guided by return maximization, agents converge to a consensus that may seem random to the human eye}~\cite{Bouchacourt2018_HowAgentsSee}. It lacks ways of anchoring its concepts in environmental, task-agnostic modalities. This is akin to the problem of \textit{symbol grounding}~\cite{Harnad1990:SymbolGrounding}. \red{Having emergent communication grounded in meanings from the environment would allow decentralized agents to learn to communicate about concepts that are shared with other agents, making learning easier and communication more efficient}~\cite{Steels1999:TalkingHeads1, Vogt2002:PhysicalGrounding, Vogt2005:EmergenceCompo}. 

Following this idea, \textit{grounding} approaches have been explored in multi-agent reinforcement learning. By linking communication with visual data~\cite{Lee2018:EmergentTranslation,Lin2021:GroundMAC}, natural language~\cite{Das2017:CoopVisDial,Havrylov2017:EmergenceLang,Gupta2021:DynamicPop,Tucker2021:DiscreteEC}, or both~\cite{Lee2019:CounteringLDrift,Lazaridou2020:MACNatLang}, agents learn to generate messages using task-agnostic concepts. \red{In other words, they learn to use concepts dictated by external modalities to transmit information efficiently, instead of searching for a consensus on their own, starting from scratch and guided only by rewards.} 
Agents may acquire "grounded" knowledge through a variety of techniques: pre-training on a supervised task~\cite{Das2017:CoopVisDial,Lee2019:CounteringLDrift,Lazaridou2020:MACNatLang,Lowe2020:S2P}, alternating between supervision and self-play~\cite{Lazaridou2017,Lowe2020:S2P}, optimizing the supervised and RL objective at the same time~\cite{Lazaridou2020:MACNatLang,Lin2021:GroundMAC,Gupta2021:DynamicPop,Tucker2021:DiscreteEC,Karten2023:IMGS-MAC}, or constructing additional rewards based on supervised models~\cite{Lee2019:CounteringLDrift}. The nature of the subsidiary tasks depends on the desired type of grounding. An autoencoding task can be added to ensure agents communicate about their observations~\cite{Lin2021:GroundMAC,Karten2023:IMGS-MAC}. To ground communication in natural language, agents can be shown examples of human-generated sentences~\cite{Das2017:CoopVisDial,Lee2018:EmergentTranslation,Gupta2021:DynamicPop,Tucker2021:DiscreteEC} or learn to generate similar outputs as pre-trained language models~\cite{Havrylov2017:EmergenceLang,Lazaridou2020:MACNatLang}. A challenge when learning to use natural language is to avoid language drift~\cite{Lazaridou2020:MACNatLang}, requiring constant supervision to prevent RL agents from forgetting the intended use of the given language~\cite{Lee2019:CounteringLDrift,Lazaridou2020:MACNatLang,Lowe2020:S2P}.
Natural language offers an efficient solution to the information bottleneck problem while allowing effortless interpretation and teaming with unknown agents (human or artificial). 

When using natural language, an obvious solution is to turn to LLMs. In addition to being extremely good for generating human-like sentences, they can also be grounded in visual and behavioral modalities~\cite{Driess2023:PaLME}. Their context window can be exploited in various ways to insert factual information or state objectives to achieve and particular behaviors to adopt~\cite{Carta2023:GLAM}. 
This is thanks to two important aspects of training the LLMs. First, the language-modeling pre-training phase shows the model of how humans formulate their reasoning in natural language. Second, the explicit instruction-following task optimized with reinforcement learning from human feedback~\cite{Christiano2017:RLHF} trains the LLM to pay close attention to what has been requested and how it should be answered.
Consequently, LLMs can be used as a basis for modeling interacting agents~\cite{Carta2023:GLAM,Driess2023:PaLME,Zhu2024:CognitiveLLMs}. Such \textit{agent-based LLMs} are given information about the environment, the task, their identity, and their role in the environment\red{, all inside an initial prompt}. Following this initialization, \red{they observe and act in the environment through visual, textual, and physical inputs and outputs~\cite{Carta2023:GLAM, Zhu2024:CognitiveLLMs}. 
Thanks to their reasoning and conversing skills, LLM agents can discuss their knowledge and intents with partners before selecting an action~\cite{Zhang2024:CoELA}.}
\red{This can even be pushed further with personas assigned to each LLM agent,} allowing a large diversity of different behaviors and offering the advantages of collective reasoning~\cite{Wu2023:AutoGen,Li2023:CAMEL,Park2023:LLMtown,Vezhnevets2023:Concordia,Perez2024:CulturalEvo}. 

LLMs offer a nice playground for multi-agent interactions. They efficiently emulate human reasoning and communication. Their built-in interactivity provides a great tool for interpretation~\cite{Wei2022:ChainOfThought} and human-agent interactions~\cite{Zhang2024:CoELA,Liu2024:HLA,Hunt2024:LLMMultiRobot}. 
However, several issues with LLMs remain and need addressing. First, embodying an LLM is a challenge requiring links to be made between language and environmental modalities (visual and behavioral). The current development of multi-modal LLMs is a step towards solving this challenge~\cite{Driess2023:PaLME}. But, these approaches often require a costly fine-tuning phase to adapt the model to its new modalities. 
A subsequent problem is the deployment of LLM-based agents on small robotic platforms, which requires engineering work to adapt to the constraints of such platforms. This is especially true for decentralized robots that must be self-sufficient and are often limited in memory and computing power. 
Furthermore, we need ways of countering the intrinsic biases present in human-generated data that LLMs inevitably reproduce~\cite{Acerbi2023:LLMsBias}. 
Lastly, the problem of hallucinations remains an important obstacle. LLMs are known for inventing information and being reluctant to admit when they are wrong~\cite{Ji2023:Hallucinations}. This can lead to issues ranging from deception to breaking the simulation, which requires more work on methods for detecting, measuring, and avoiding these hallucinations. 
While these issues can, and will certainly be addressed, this reminds us that other solutions using smaller models also work and might be preferable in many situations.

\red{To conclude, we see that many approaches exist for teaching decentralized agents to communicate about high-dimensional environmental features. They rely on languages that select information to communicate more efficiently. These languages abstract physical elements of the world by grounding symbols in environmental features, allowing the establishment of conventions on how information should be transmitted. Different degrees of physical abstraction may serve different purposes. A group of agents specialized in a single task may be content with low physically-abstracted differentiable emergent communication learned from task reward. On the other hand, established concepts and grammatical rules provide the tools to generalize acquired knowledge, compose new ideas from fundamental language blocks, and communicate with unknown partners. Thus, higher physical abstraction, found in natural languages, is better fit to handle more general settings. }
%


\section{Conclusion}
\label{sec-conclusion}

We explored how communication through signaling can be crucial in enhancing coordination within robot swarms operating under the Decentralized Learning and Execution paradigm. We proposed a structured framework to classify existing and future signaling methods, covering a wide range of information \red{selection} levels and physical abstractions. Throughout the paper, we advocate that swarm robotics with distributed online learning capabilities offer unique challenges, for which communication can play a positive role, but to which communication is also subject. The key messages of our paper are summarized hereafter.

\red{\textbf{A path towards complex communication strategies.}} Earlier works in swarm robotics were closely inspired by social insects. The current state-of-the-art in swarm robotics now shows a great variety of applications and robotics setups, including dense to sparse swarms with homogeneous or heterogeneous robots. To account for the fast-paced advances in hardware and software, it is important to keep in mind that swarm robotics is about the relation between microscopic interactions and macroscopic organization, which remains valid even if powerful computation and signaling capabilities are available. A practical consequence we envision is the advent of robots using large language models, composing a society of embodied agents with human-like signaling capabilities that are still bound by environmental contingencies (e.g., local communication only, complex physical interactions). 
\red{Beyond the anticipated gains in performance, incorporating human language-like capabilities can offer valuable benefits with respect to explainability and human-robot interaction through the use of a shared language.} 

\red{\textbf{Decentralized learning and adaptive dynamics in Swarm Robotics present a unique challenge.}}
Addressing the problem of distributed credit assignment is a well-known challenge in multi-agent systems. However, conducting learning in a decentralized and online fashion adds another layer of complexity, especially when policy parameters hop from one robot to the next. A consequence is that nearby robots can share similar parameters, which can indirectly cause either altruistic or competitive behaviors depending on the degree of relation between individuals (see Section~\ref{sec-dynamics}). Differing from natural systems where the population may grow, the fixed size of a robot swarm impacts where competition occurs: robots are mere resources for which policy parameters are competing, rather than the opposite. Exploring the long-term adaptive dynamics of behavioral strategies (in which signaling is included) in dynamic and unpredictable environments will be critical for developing adaptive and resilient swarm systems. \red{This opens up an exciting avenue, requiring an interdisciplinary research effort, integrating expertise from fields such as evolutionary game theory~\cite{fudenberg1998learning}, collective decision-making~\cite{Nisan2007agt}, evolutionary dynamics~\cite{nowak2006evolutionary}, sociophysics~\cite{sen2014sociophysics,galam1982sociophysics}, physics of active matter~\cite{tailleur2022active},  evolutionary computation~\cite{eiben2015introduction} and machine learning~\cite{sutton2018reinforcement}.}
\red{
We conclude with a list of take-home messages, targeting the three communities we believe will be at the center of this coming revolution:

\begin{itemize}
\item \textbf{Researchers in Swarm Robotics}: simple robots are not inherently "simple". What matters is the emergence of complex behaviors from microscopic interactions. Whether you work with large or small robots, dense or sparse populations, or few or many robots, all are welcome under the broad aim of continual learning in swarm systems.

\item \textbf{Researchers in Machine Learning}: this is all about embodiment. Swarm robotics introduces a unique category of machine learning problems with elements of "social" learning across physically embodied agents. Anchoring language models in physical systems brings new challenges and capabilities in distributed, online learning.

\item \textbf{Researchers in Complex Systems}: swarm robotics provides a controllable model for exploring active matter, sociophysics models, reaction-diffusion, and diffusion biophysics processes. Swarm robotics offers an experimental platform for addressing fundamental questions about adaptive collective systems.
\end{itemize}
}

We believe that decentralized learning and execution will inevitably become more prominent in swarm robotics, with signaling playing a fundamental role. We intend for this paper to serve as a milestone in shaping the future of this field by providing a framework to understand the complexities and potentials of swarm robotics, where local interactions drive continuously learning embodied agents equipped with complex signaling mechanisms.


\section*{Acknowledgment}

This work was supported by the SSR project funded by the Agence Nationale pour la Recherche under Grant No ANR-24-CE33-7791.


\enlargethispage{20pt}



\vskip2pc

\bibliography{nicolas,maxime,leo}

\begin{thebibliography}{100}

\bibitem{beni1993swarm}
G~Beni and J~Wang.
\newblock Swarm intelligence in cellular robotic systems.
\newblock In {\em NATO ASI}. 1993.

\bibitem{dudek1993taxonomy}
Gregory Dudek, Michael Jenkin, Evangelos Milios, and David Wilkes.
\newblock A taxonomy for swarm robots.
\newblock In {\em Proceedings of 1993 IEEE/RSJ International Conference on Intelligent Robots and Systems (IROS'93)}, volume~1, pages 441--447. IEEE, 1993.

\bibitem{Brambilla2013}
Manuele Brambilla, Eliseo Ferrante, Mauro Birattari, and Marco Dorigo.
\newblock {Swarm robotics : A review from the swarm engineering perspective}.
\newblock {\em Swarm Intelligence}, 7(1):1--41, 2013.

\bibitem{Hamman2018}
Heiko Hamann.
\newblock {\em Swarm Robotics - {A} Formal Approach}.
\newblock Springer, 2018.

\bibitem{dorigo2020reflections}
M~Dorigo, G~Theraulaz, and V~Trianni.
\newblock Reflections on the future of swarm robotics.
\newblock {\em Science Robotics}, 2020.

\bibitem{floreano2021individual}
D~Floreano and H~Lipson.
\newblock From individual robots to robot societies, 2021.

\bibitem{watson2002embodied}
RA~Watson, SG~Ficici, and JB~Pollack.
\newblock Embodied evolution: Distributing an evolutionary algorithm in a population of robots.
\newblock {\em Robotics and Autonomous Systems}, 2002.

\bibitem{heinerman2015evolution}
Jacqueline Heinerman, Massimiliano Rango, and Agoston~Endre Eiben.
\newblock Evolution, individual learning, and social learning in a swarm of real robots.
\newblock In {\em 2015 IEEE symposium series on computational intelligence}, pages 1055--1062. IEEE, 2015.

\bibitem{bredeche2022rspt}
N~Bredeche and N~Fontbonne.
\newblock Social learning in swarm robotics.
\newblock {\em Philosophical Transactions of the Royal Society B}, 2022.

\bibitem{bredeche2018ee}
N~Bredeche, E~Haasdijk, and A~Prieto.
\newblock {Embodied evolution in collective robotics: A review}.
\newblock {\em Frontiers in Robotics and AI}, 2018.

\bibitem{zhang2018fully}
Kaiqing Zhang, Zhuoran Yang, Han Liu, Tong Zhang, and Tamer Basar.
\newblock Fully decentralized multi-agent reinforcement learning with networked agents.
\newblock In {\em International Conference on Machine Learning}, pages 5872--5881. PMLR, 2018.

\bibitem{lyu2021contrasting}
Xueguang Lyu, Yuchen Xiao, Brett Daley, and Christopher Amato.
\newblock Contrasting centralized and decentralized critics in multi-agent reinforcement learning.
\newblock {\em arXiv preprint arXiv:2102.04402}, 2021.

\bibitem{ficici1999embodied}
Sevan~G Ficici, Richard~A Watson, and Jordan~B Pollack.
\newblock Embodied evolution: A response to challenges in evolutionary robotics.
\newblock In {\em Proceedings of the eighth European workshop on learning robots}, pages 14--22. Citeseer, 1999.

\bibitem{bredeche2010ppsn}
Nicolas Bredeche and Jean-Marc Montanier.
\newblock Environment-driven embodied evolution in a population of autonomous agents.
\newblock In {\em International Conference on Parallel Problem Solving from Nature}, pages 290--299. Springer, 2010.

\bibitem{schranz2020swarmapplis}
Melanie Schranz, Martina Umlauft, Micha Sende, and Wilfried Elmenreich.
\newblock Swarm robotic behaviors and current applications.
\newblock {\em Frontiers in Robotics and AI}, 7:36, 2020.

\bibitem{Nisan2007agt}
Noam Nisan, Tim Roughgarden, \'Eva Tardos, and Vijay~V. Vazirani.
\newblock {\em Algorithmic Game Theory}.
\newblock Cambridge University Press, New York, NY, USA, 2007.

\bibitem{oroojlooy2023review}
Afshin Oroojlooy and Davood Hajinezhad.
\newblock A review of cooperative multi-agent deep reinforcement learning.
\newblock {\em Applied Intelligence}, 53(11):13677--13722, 2023.

\bibitem{wolpert1999introduction}
David~H Wolpert and Kagan Tumer.
\newblock An introduction to collective intelligence.
\newblock {\em arXiv preprint cs/9908014}, 1999.

\bibitem{stone2010ad}
Peter Stone, Gal Kaminka, Sarit Kraus, and Jeffrey Rosenschein.
\newblock Ad hoc autonomous agent teams: Collaboration without pre-coordination.
\newblock In {\em Proceedings of the AAAI Conference on Artificial Intelligence}, volume~24, pages 1504--1509, 2010.

\bibitem{marden2018game}
Jason~R Marden and Jeff~S Shamma.
\newblock Game theory and control.
\newblock {\em Annual review of control, robotics, and autonomous systems}, 1(1):105--134, 2018.

\bibitem{ecoffet2021nothing}
P~Ecoffet, N~Bredeche, and JB~Andr{\'e}.
\newblock Nothing better to do? environment quality and the evolution of cooperation by partner choice.
\newblock {\em Journal of Theoretical Biology}, 2021.

\bibitem{Shapley1953}
Lloyd~S Shapley.
\newblock {A Value for n-person Games}.
\newblock {\em Annals of Mathematical Studies}, 28:307–317, 1953.

\bibitem{shoham2008multiagent}
Yoav Shoham and Kevin Leyton-Brown.
\newblock {\em Multiagent systems: Algorithmic, game-theoretic, and logical foundations}.
\newblock Cambridge University Press, 2008.

\bibitem{wooldridge2009introduction}
Michael Wooldridge.
\newblock {\em An introduction to multiagent systems}.
\newblock John wiley \& sons, 2009.

\bibitem{Trianni2006}
Vito Trianni and Marco Dorigo.
\newblock Self-organisation and communication in groups of simulated and physical robots.
\newblock {\em Biological Cybernetics}, 95(3):213--231, 2006.

\bibitem{waibel2009genetic}
Markus Waibel, Laurent Keller, and Dario Floreano.
\newblock Genetic team composition and level of selection in the evolution of cooperation.
\newblock {\em IEEE transactions on Evolutionary Computation}, 13(3):648--660, 2009.

\bibitem{Wolpert2000OptimalWL}
D.~Wolpert, Kagan Tumer, and K.~Swanson.
\newblock Optimal wonderful life utility functions in multi-agent systems.
\newblock 2000.

\bibitem{kolpaczki2024approximating}
Patrick Kolpaczki, Viktor Bengs, Maximilian Muschalik, and Eyke H{\"u}llermeier.
\newblock Approximating the shapley value without marginal contributions.
\newblock In {\em Proceedings of the AAAI Conference on Artificial Intelligence}, volume~38, pages 13246--13255, 2024.

\bibitem{wang2022cooperative}
Jianrui Wang, Yitian Hong, Jiali Wang, Jiapeng Xu, Yang Tang, Qing-Long Han, and J{\"u}rgen Kurths.
\newblock Cooperative and competitive multi-agent systems: From optimization to games.
\newblock {\em IEEE/CAA Journal of Automatica Sinica}, 9(5):763--783, 2022.

\bibitem{gronauer2022multi}
Sven Gronauer and Klaus Diepold.
\newblock Multi-agent deep reinforcement learning: a survey.
\newblock {\em Artificial Intelligence Review}, 55(2):895--943, 2022.

\bibitem{dawkins2016selfish}
Richard Dawkins.
\newblock {\em The selfish gene}.
\newblock Oxford university press, 2016.

\bibitem{west2007social}
Stuart~A West, Ashleigh~S Griffin, and Andy Gardner.
\newblock Social semantics: altruism, cooperation, mutualism, strong reciprocity and group selection.
\newblock {\em Journal of evolutionary biology}, 20(2):415--432, 2007.

\bibitem{hamilton1964}
William~D Hamilton.
\newblock The genetical evolution of social behaviour. ii.
\newblock {\em Journal of theoretical biology}, 7(1):17--52, 1964.

\bibitem{richerson2008not}
Peter~J Richerson and Robert Boyd.
\newblock {\em Not by genes alone: How culture transformed human evolution}.
\newblock University of Chicago press, 2008.

\bibitem{smith2010communication}
Eric~Alden Smith.
\newblock Communication and collective action: language and the evolution of human cooperation.
\newblock {\em Evolution and human behavior}, 31(4):231--245, 2010.

\bibitem{montanier2011surviving}
Jean-Marc Montanier and Nicolas Bredeche.
\newblock Surviving the tragedy of commons: emergence of altruism in a population of evolving autonomous agents.
\newblock In {\em European conference on artificial life}, 2011.

\bibitem{waibel2011quantitative}
Markus Waibel, Dario Floreano, and Laurent Keller.
\newblock A quantitative test of hamilton's rule for the evolution of altruism.
\newblock {\em PLoS biology}, 9(5):e1000615, 2011.

\bibitem{zhang2021decentralized}
Kaiqing Zhang, Zhuoran Yang, and Tamer Ba{\c{s}}ar.
\newblock Decentralized multi-agent reinforcement learning with networked agents: Recent advances.
\newblock {\em Frontiers of Information Technology \& Electronic Engineering}, 22(6):802--814, 2021.

\bibitem{zimmer2021learning}
Matthieu Zimmer, Claire Glanois, Umer Siddique, and Paul Weng.
\newblock Learning fair policies in decentralized cooperative multi-agent reinforcement learning.
\newblock In {\em International Conference on Machine Learning}, pages 12967--12978. PMLR, 2021.

\bibitem{foerster2018counterfactual}
Jakob Foerster, Gregory Farquhar, Triantafyllos Afouras, Nantas Nardelli, and Shimon Whiteson.
\newblock Counterfactual multi-agent policy gradients.
\newblock In {\em Proceedings of the AAAI conference on artificial intelligence}, volume~32, 2018.

\bibitem{wischmann2012historical}
Steffen Wischmann, Dario Floreano, and Laurent Keller.
\newblock Historical contingency affects signaling strategies and competitive abilities in evolving populations of simulated robots.
\newblock {\em Proceedings of the National Academy of Sciences}, 109(3):864--868, 2012.

\bibitem{floreano2007evolutionary}
Dario Floreano, Sara Mitri, St{\'e}phane Magnenat, and Laurent Keller.
\newblock Evolutionary conditions for the emergence of communication in robots.
\newblock {\em Current biology}, 17(6):514--519, 2007.

\bibitem{reynolds1987flocks}
CW~Reynolds.
\newblock Flocks, herds and schools: A distributed behavioral model.
\newblock In {\em SIGGRAPH}, 1987.

\bibitem{vicsek1995novel}
T~Vicsek, A~Czir{\'o}k, E~Ben-Jacob, I~Cohen, and O~Shochet.
\newblock Novel type of phase transition in a system of self-driven particles.
\newblock {\em Physical review letters}, 1995.

\bibitem{smith2003animal}
John~Maynard Smith and David Harper.
\newblock {\em Animal signals}.
\newblock Oxford University Press, 2003.

\bibitem{Berghe2018socroblg}
Rianne van~den Berghe, Josje Verhagen, Ora Oudgenoeg-Paz, Sanne van~der Ven, and Paul Leseman.
\newblock Social robots for language learning: A review.
\newblock {\em Review of Educational Research}, 89(2):259--295, 2019.

\bibitem{bonabeau2000inspiration}
Eric Bonabeau, Marco Dorigo, and Guy Theraulaz.
\newblock Inspiration for optimization from social insect behaviour.
\newblock {\em Nature}, 406(6791):39--42, 2000.

\bibitem{detrain2008collective}
C~Detrain and JL~Deneubourg.
\newblock Collective decision-making and foraging patterns in ants and honeybees.
\newblock {\em Advances in insect physiology}, 2008.

\bibitem{bonabeau1999swarm}
Eric Bonabeau, Marco Dorigo, and Guy Theraulaz.
\newblock {\em Swarm intelligence: from natural to artificial systems}.
\newblock Oxford university press, 1999.

\bibitem{campo2010artificial}
Alexandre Campo, {\'A}lvaro Guti{\'e}rrez, Shervin Nouyan, Carlo Pinciroli, Valentin Longchamp, Simon Garnier, and Marco Dorigo.
\newblock Artificial pheromone for path selection by a foraging swarm of robots.
\newblock {\em Biological cybernetics}, 103:339--352, 2010.

\bibitem{carrasco2011visual}
Marisa Carrasco.
\newblock Visual attention: The past 25 years.
\newblock {\em Vision research}, 51(13):1484--1525, 2011.

\bibitem{bellman1966dynamic}
Richard Bellman.
\newblock Dynamic programming.
\newblock {\em science}, 153(3731):34--37, 1966.

\bibitem{cazenille2024hearing}
L~Cazenille, N~Lobato-Dauzier, A~Loi, M~Ito, O~Marchal, N~Aubert-Kato, N~Bredeche, and AJ~Genot.
\newblock Hearing the shape of an arena with spectral swarm robotics.
\newblock {\em arXiv:2403.17147}, 2024.

\bibitem{gerganov_llama_cpp}
Georgi Gerganov.
\newblock llama.cpp.
\newblock \url{https://github.com/ggerganov/llama.cpp}, 2023.
\newblock Accessed: 2024-05-30.

\bibitem{ji2023signal}
Peng Ji, Jiachen Ye, Yu~Mu, Wei Lin, Yang Tian, Chittaranjan Hens, Matja{\v{z}} Perc, Yang Tang, Jie Sun, and J{\"u}rgen Kurths.
\newblock Signal propagation in complex networks.
\newblock {\em Physics reports}, 1017:1--96, 2023.

\bibitem{taga2003chemical}
Michiko~E Taga and Bonnie~L Bassler.
\newblock Chemical communication among bacteria.
\newblock {\em Proceedings of the National Academy of Sciences}, 100(suppl\_2):14549--14554, 2003.

\bibitem{larue2015acoustic}
Kelly~M LaRue, Jan Clemens, Gordon~J Berman, and Mala Murthy.
\newblock Acoustic duetting in drosophila virilis relies on the integration of auditory and tactile signals.
\newblock {\em Elife}, 4:e07277, 2015.

\bibitem{bairos2019novel}
Kevin~R Bairos-Novak, Maud~CO Ferrari, and Douglas~P Chivers.
\newblock A novel alarm signal in aquatic prey: familiar minnows coordinate group defences against predators through chemical disturbance cues.
\newblock {\em Journal of Animal Ecology}, 88(9):1281--1290, 2019.

\bibitem{marques2009firefly}
Simone~M Marques and Joaquim~CG Esteves~da Silva.
\newblock Firefly bioluminescence: a mechanistic approach of luciferase catalyzed reactions.
\newblock {\em IUBMB life}, 61(1):6--17, 2009.

\bibitem{hopkins1974electric}
Carl~D Hopkins.
\newblock Electric communication in fish.
\newblock {\em American Scientist}, 62(4):426--437, 1974.

\bibitem{eriksson1986male}
Dag Eriksson and Lars Wallin.
\newblock Male bird song attracts females—a field experiment.
\newblock {\em Behavioral Ecology and Sociobiology}, 19:297--299, 1986.

\bibitem{searcy2006bird}
William~A Searcy, Rindy~C Anderson, and Stephen Nowicki.
\newblock Bird song as a signal of aggressive intent.
\newblock {\em Behavioral Ecology and Sociobiology}, 60:234--241, 2006.

\bibitem{rodrigues2015overcoming}
Tiago Rodrigues, Miguel Duarte, Margarida Figueir{\'o}, Vasco Costa, Sancho~Moura Oliveira, and Anders~Lyhne Christensen.
\newblock Overcoming limited onboard sensing in swarm robotics through local communication.
\newblock In {\em Transactions on Computational Collective Intelligence XX}, pages 201--223. Springer, 2015.

\bibitem{talamali2021less}
Mohamed~S Talamali, Arindam Saha, James~AR Marshall, and Andreagiovanni Reina.
\newblock When less is more: Robot swarms adapt better to changes with constrained communication.
\newblock {\em Science Robotics}, 6(56):eabf1416, 2021.

\bibitem{mcguire2019minimal}
KN~McGuire, Christophe De~Wagter, Karl Tuyls, HJ~Kappen, and Guido~CHE de~Croon.
\newblock Minimal navigation solution for a swarm of tiny flying robots to explore an unknown environment.
\newblock {\em Science Robotics}, 4(35):eaaw9710, 2019.

\bibitem{hafnaoui2019timing}
Imane Hafnaoui, Gabriela Nicolescu, and Giovanni Beltrame.
\newblock Timing information propagation in interactive networks.
\newblock {\em Scientific Reports}, 9(1):4442, 2019.

\bibitem{crank1979mathematics}
John Crank.
\newblock {\em The mathematics of diffusion}.
\newblock Oxford university press, 1979.

\bibitem{shanks2001modeling}
Niall Shanks.
\newblock Modeling biological systems: the belousov--zhabotinsky reaction.
\newblock {\em Foundations of Chemistry}, 3(1):33--53, 2001.

\bibitem{turing1990chemical}
AM~Turing.
\newblock The chemical basis of morphogenesis.
\newblock {\em Bulletin of mathematical biology}, 52(1-2):153--197, 1990.

\bibitem{arai1993information}
Tamio Arai, Eiichi Yoshida, and Jun Ota.
\newblock Information diffusion by local communication of multiple mobile robots.
\newblock In {\em Proceedings of IEEE Systems Man and Cybernetics Conference-SMC}, volume~4, pages 535--540. IEEE, 1993.

\bibitem{murata2022molecular}
Satoshi Murata.
\newblock {\em Molecular Robotics: An Introduction}.
\newblock Springer, 2022.

\bibitem{nummelin2020robotic}
Sami Nummelin, Boxuan Shen, Petteri Piskunen, Qing Liu, Mauri~A Kostiainen, and Veikko Linko.
\newblock Robotic {DNA} nanostructures.
\newblock {\em ACS Synthetic Biology}, 9(8):1923--1940, 2020.

\bibitem{douglas2012logic}
Shawn~M Douglas, Ido Bachelet, and George~M Church.
\newblock A logic-gated nanorobot for targeted transport of molecular payloads.
\newblock {\em Science}, 335(6070):831--834, 2012.

\bibitem{torelli2014dna}
Emanuela Torelli, Monica Marini, Sabrina Palmano, Luca Piantanida, Cesare Polano, Alice Scarpellini, Marco Lazzarino, and Giuseppe Firrao.
\newblock A {DNA} origami nanorobot controlled by nucleic acid hybridization.
\newblock {\em Small}, 10(14):2918--2926, 2014.

\bibitem{kuzuya2014nanomechanical}
Akinori Kuzuya and Yuichi Ohya.
\newblock Nanomechanical molecular devices made of {DNA} origami.
\newblock {\em Accounts of chemical research}, 47(6):1742--1749, 2014.

\bibitem{amir2015folding}
Yaniv Amir, Almogit Abu-Horowitz, and Ido Bachelet.
\newblock Folding and characterization of a bio-responsive robot from {DNA} origami.
\newblock {\em JoVE (Journal of Visualized Experiments)}, (106):e51272, 2015.

\bibitem{kaminka2017molecular}
Gal~A Kaminka, Rachel Spokoini-Stern, Yaniv Amir, Noa Agmon, and Ido Bachelet.
\newblock Molecular robots obeying asimov's three laws of robotics.
\newblock {\em Artificial life}, 23(3):343--350, 2017.

\bibitem{daljit2018switchable}
Jasleen~Kaur Daljit~Singh, Minh~Tri Luu, Ali Abbas, and Shelley~FJ Wickham.
\newblock Switchable {DNA}-origami nanostructures that respond to their environment and their applications.
\newblock {\em Biophysical reviews}, 10(5):1283--1293, 2018.

\bibitem{li2018dna}
Suping Li, Qiao Jiang, Shaoli Liu, Yinlong Zhang, Yanhua Tian, Chen Song, Jing Wang, Yiguo Zou, Gregory~J Anderson, Jing-Yan Han, et~al.
\newblock A {DNA} nanorobot functions as a cancer therapeutic in response to a molecular trigger in vivo.
\newblock {\em Nature biotechnology}, 36(3):258--264, 2018.

\bibitem{wickham2012dna}
Shelley~FJ Wickham, Jonathan Bath, Yousuke Katsuda, Masayuki Endo, Kumi Hidaka, Hiroshi Sugiyama, and Andrew~J Turberfield.
\newblock A {DNA}-based molecular motor that can navigate a network of tracks.
\newblock {\em Nature nanotechnology}, 7(3):169--173, 2012.

\bibitem{thubagere2017cargo}
Anupama~J Thubagere, Wei Li, Robert~F Johnson, Zibo Chen, Shayan Doroudi, Yae~Lim Lee, Gregory Izatt, Sarah Wittman, Niranjan Srinivas, Damien Woods, et~al.
\newblock A cargo-sorting {DNA} robot.
\newblock {\em Science}, 357(6356):eaan6558, 2017.

\bibitem{gines2017microscopic}
Guillaume Gines, AS~Zadorin, J-C Galas, Teruo Fujii, A~Estevez-Torres, and Y~Rondelez.
\newblock Microscopic agents programmed by {DNA} circuits.
\newblock {\em Nature nanotechnology}, 12(4):351--359, 2017.

\bibitem{aubert2017evolutionary}
Nathanael Aubert-Kato, Charles Fosseprez, Guillaume Gines, Ibuki Kawamata, Huy Dinh, Leo Cazenille, Andre Estevez-Tores, Masami Hagiya, Yannick Rondelez, and Nicolas Bredeche.
\newblock Evolutionary optimization of self-assembly in a swarm of bio-micro-robots.
\newblock In {\em Proceedings of the Genetic and Evolutionary Computation Conference}, pages 59--66, 2017.

\bibitem{cazenille2019exploring}
Leo Cazenille, Nicolas Bredeche, and Nathanael Aubert-Kato.
\newblock Exploring self-assembling behaviors in a swarm of bio-micro-robots using surrogate-assisted map-elites.
\newblock In {\em 2019 IEEE Symposium Series on Computational Intelligence (SSCI)}, pages 238--246. IEEE, 2019.

\bibitem{ziepke2022multi}
Alexander Ziepke, Ivan Maryshev, Igor~S Aranson, and Erwin Frey.
\newblock Multi-scale organization in communicating active matter.
\newblock {\em Nature communications}, 13(1):6727, 2022.

\bibitem{wang2024swarm}
Yibin Wang, Hui Chen, Leiming Xie, Jinbo Liu, Li~Zhang, and Jiangfan Yu.
\newblock Swarm autonomy: From agent functionalization to machine intelligence.
\newblock {\em Advanced Materials}, page 2312956, 2024.

\bibitem{grauer2024optimizing}
Jens Grauer, Fabian~Jan Schwarzendahl, Hartmut L{\"o}wen, and Benno Liebchen.
\newblock Optimizing collective behavior of communicating active particles with machine learning.
\newblock {\em Machine Learning: Science and Technology}, 5(1):015014, 2024.

\bibitem{lavergne2019group}
Fran{\c{c}}ois~A Lavergne, Hugo Wendehenne, Tobias B{\"a}uerle, and Clemens Bechinger.
\newblock Group formation and cohesion of active particles with visual perception--dependent motility.
\newblock {\em Science}, 364(6435):70--74, 2019.

\bibitem{keya2018dna}
Jakia~Jannat Keya, Ryuhei Suzuki, Arif Md~Rashedul Kabir, Daisuke Inoue, Hiroyuki Asanuma, Kazuki Sada, Henry Hess, Akinori Kuzuya, and Akira Kakugo.
\newblock {DNA}-assisted swarm control in a biomolecular motor system.
\newblock {\em Nature communications}, 9(1):453, 2018.

\bibitem{akter2022cooperative}
Mousumi Akter, JJ~Keya, K~Kayano, AMR Kabir, Daisuke Inoue, Henry Hess, K~Sada, Akinori Kuzuya, Hiroyuki Asanuma, and Akira Kakugo.
\newblock Cooperative cargo transportation by a swarm of molecular machines.
\newblock {\em Science Robotics}, 7(65):eabm0677, 2022.

\bibitem{aubert2023collective}
Nathanael Aubert-Kato, Geoff Nitschke, Ibuki Kawamata, and Akira Kakugo.
\newblock Collective cargo transport and sorting with molecular swarms.
\newblock In {\em Artificial Life Conference Proceedings 35}, volume 2023, page~90. MIT Press One Rogers Street, Cambridge, MA 02142-1209, USA journals-info~…, 2023.

\bibitem{slavkov2018morphogenesis}
I~Slavkov, D~Carrillo-Zapata, N~Carranza, X~Diego, F~Jansson, J~Kaandorp, S~Hauert, and J~Sharpe.
\newblock Morphogenesis in robot swarms.
\newblock {\em Science Robotics}, 2018.

\bibitem{rubenstein2014programmable}
M~Rubenstein, A~Cornejo, and R~Nagpal.
\newblock Programmable self-assembly in a thousand-robot swarm.
\newblock {\em Science}, 2014.

\bibitem{gauci2017error}
M~Gauci, ME~Ortiz, M~Rubenstein, and R~Nagpal.
\newblock Error cascades in collective behavior: a case study of the gradient algorithm on 1000 physical agents.
\newblock In {\em Proceedings of the 16th Conference on Autonomous Agents and MultiAgent Systems}, pages 1404--1412, 2017.

\bibitem{wang2020fast}
H~Wang and M~Rubenstein.
\newblock A fast, accurate, and scalable probabilistic sample-based approach for counting swarm size.
\newblock In {\em 2020 IEEE International Conference on Robotics and Automation (ICRA)}, pages 7180--7185. IEEE, 2020.

\bibitem{busoniu2008comprehensive}
Lucian Busoniu, Robert Babuska, and Bart De~Schutter.
\newblock A comprehensive survey of multiagent reinforcement learning.
\newblock {\em IEEE Transactions on Systems, Man, and Cybernetics, Part C (Applications and Reviews)}, 38(2):156--172, 2008.

\bibitem{saad2011numerical}
Yousef Saad.
\newblock {\em Numerical methods for large eigenvalue problems: revised edition}.
\newblock SIAM, 2011.

\bibitem{oppenheim1999discrete}
Alan~V Oppenheim.
\newblock {\em Discrete-time signal processing}.
\newblock Pearson Education India, 1999.

\bibitem{Ohmer2022:MutualInfluence}
Xenia Ohmer, Michael Marino, Michael Franke, and Peter König.
\newblock Mutual influence between language and perception in multi-agent communication games.
\newblock {\em PLOS Computational Biology}, 18(10):1--28, 10 2022.

\bibitem{Tishby1999:IB}
Naftali Tishby, Fernando~C. Pereira, and William Bialek.
\newblock The information bottleneck method.
\newblock In {\em Proceedings of the 37-th Annual Allerton Conference on Communication, Control and Computing}, pages 368--377, 1999.

\bibitem{Kirby2015_CompressionComm}
Simon Kirby, Monica Tamariz, Hannah Cornish, and Kenny Smith.
\newblock Compression and communication in the cultural evolution of linguistic structure.
\newblock {\em Cognition}, 141:87--102, August 2015.

\bibitem{Zaslavsky2018:ColorNamingEvo}
Noga Zaslavsky, Charles Kemp, Terry Regier, and Naftali Tishby.
\newblock Efficient compression in color naming and its evolution.
\newblock {\em Proceedings of the National Academy of Sciences}, 115(31):7937--7942, July 2018.

\bibitem{Steels1999:TalkingHeads1}
Luc Steels.
\newblock {\em The Talking Heads Experiment. Volume I. Words and Meanings}.
\newblock 1999.

\bibitem{Kirby2001:ILM}
S.~Kirby.
\newblock Spontaneous evolution of linguistic structure-an iterated learning model of the emergence of regularity and irregularity.
\newblock {\em IEEE Transactions on Evolutionary Computation}, 5(2):102--110, April 2001.

\bibitem{Smith2003:IteratedLearning}
Kenny Smith, Simon Kirby, and Henry Brighton.
\newblock Iterated learning: A framework for the emergence of language.
\newblock {\em Artificial Life}, 9(4):371--386, October 2003.

\bibitem{Vogt2005:EmergenceCompo}
Paul Vogt.
\newblock The emergence of compositional structures in perceptually grounded language games.
\newblock {\em Artificial Intelligence}, 167(1–2):206--242, September 2005.

\bibitem{Perfors2014_WorldShapeLang}
Andrew Perfors and Daniel~J. Navarro.
\newblock Language evolution can be shaped by the structure of the world.
\newblock {\em Cognitive Science}, 38(4):775--793, January 2014.

\bibitem{Christiansen2008_BrainShapeLang}
Morten~H. Christiansen and Nick Chater.
\newblock Language as shaped by the brain.
\newblock {\em Behavioral and Brain Sciences}, 31(5):489--509, October 2008.

\bibitem{Wellens2008:FlexibleWordMeaning}
Peter Wellens, Martin Loetzsch, and Luc Steels.
\newblock Flexible word meaning in embodied agents.
\newblock {\em Connection Science}, 20(2–3):173--191, September 2008.

\bibitem{Beuls2013:GrammaticalAgreement}
Katrien Beuls and Luc Steels.
\newblock Agent-based models of strategies for the emergence and evolution of grammatical agreement.
\newblock {\em PLoS ONE}, 8(3):e58960, March 2013.

\bibitem{Ekila2024:LinguisticConventions}
J\'{e}r\^{o}me Botoko~Ekila.
\newblock Emergence of linguistic conventions in multi-agent systems through situated communicative interactions.
\newblock In {\em Proceedings of the 23rd International Conference on Autonomous Agents and Multiagent Systems}, AAMAS '24, page 2725–2727, Richland, SC, 2024. International Foundation for Autonomous Agents and Multiagent Systems.

\bibitem{Wagner2003:EC}
Kyle Wagner, James~A. Reggia, Juan Uriagereka, and Gerald~S. Wilkinson.
\newblock Progress in the simulation of emergent communication and language.
\newblock {\em Adaptive Behavior}, 11(1):37--69, 2003.

\bibitem{Zhu2024_MACSurvey}
Changxi Zhu, Mehdi Dastani, and Shihan Wang.
\newblock A survey of multi-agent deep reinforcement learning with communication.
\newblock {\em Autonomous Agents and Multi-Agent Systems}, 38(1), January 2024.

\bibitem{Sukhbaatar2016:CommNet}
Sainbayar Sukhbaatar, Arthur Szlam, and Rob Fergus.
\newblock Learning multiagent communication with backpropagation.
\newblock In {\em Advances in Neural Information Processing Systems, pp. 2244–2252}, 2016.

\bibitem{Foerster2016:DIAL}
Jakob~N. Foerster, Yannis~M. Assael, Nando de~Freitas, and Shimon Whiteson.
\newblock Learning to communicate with deep multi-agent reinforcement learning.
\newblock In {\em Proceedings of the 30th International Conference on Neural Information Processing Systems}, NIPS'16, page 2145–2153, Red Hook, NY, USA, 2016. Curran Associates Inc.

\bibitem{Peng2017:BiCNet}
Peng Peng, Ying Wen, Yaodong Yang, Quan Yuan, Zhenkun Tang, Haitao Long, and Jun Wang.
\newblock Multiagent bidirectionally-coordinated nets: Emergence of human-level coordination in learning to play starcraft combat games.
\newblock 2017.

\bibitem{Wang2022:FCMNet}
Yutong Wang and Guillaume Sartoretti.
\newblock Fcmnet: Full communication memory net for team-level cooperation in multi-agent systems.
\newblock In {\em Proceedings of the 21st International Conference on Autonomous Agents and Multiagent Systems}, AAMAS '22, page 1355–1363, Richland, SC, 2022. International Foundation for Autonomous Agents and Multiagent Systems.

\bibitem{Hoshen2017:VAIN}
Yedid Hoshen.
\newblock Vain: Attentional multi-agent predictive modeling.
\newblock In {\em Proceedings of the 31st International Conference on Neural Information Processing Systems}, NIPS'17, page 2698–2708, Red Hook, NY, USA, 2017. Curran Associates Inc.

\bibitem{Jiang2018:ATOC}
Jiechuan Jiang and Zongqing Lu.
\newblock Learning attentional communication for multi-agent cooperation.
\newblock In S.~Bengio, H.~Wallach, H.~Larochelle, K.~Grauman, N.~Cesa-Bianchi, and R.~Garnett, editors, {\em Advances in Neural Information Processing Systems}, volume~31. Curran Associates, Inc., 2018.

\bibitem{Das2019:TarMAC}
Abhishek Das, Th{\'e}ophile Gervet, Joshua Romoff, Dhruv Batra, Devi Parikh, Mike Rabbat, and Joelle Pineau.
\newblock {T}ar{MAC}: Targeted multi-agent communication.
\newblock In Kamalika Chaudhuri and Ruslan Salakhutdinov, editors, {\em Proceedings of the 36th International Conference on Machine Learning}, volume~97 of {\em Proceedings of Machine Learning Research}, pages 1538--1546. PMLR, 09--15 Jun 2019.

\bibitem{Singh2019:IC3Net}
Amanpreet Singh, Tushar Jain, and Sainbayar Sukhbaatar.
\newblock Learning when to communicate at scale in multiagent cooperative and competitive tasks.
\newblock In {\em International Conference on Learning Representations}, 2019.

\bibitem{Zhang2019:VBC}
Sai~Qian Zhang, Qi~Zhang, and Jieyu Lin.
\newblock Efficient communication in multi-agent reinforcement learning via variance based control.
\newblock In H.~Wallach, H.~Larochelle, A.~Beygelzimer, F.~d\textquotesingle Alch\'{e}-Buc, E.~Fox, and R.~Garnett, editors, {\em Advances in Neural Information Processing Systems}, volume~32. Curran Associates, Inc., 2019.

\bibitem{Wang2020:IMAC}
Rundong Wang, Xu~He, Runsheng Yu, Wei Qiu, Bo~An, and Zinovi Rabinovich.
\newblock Learning efficient multi-agent communication: An information bottleneck approach.
\newblock In Hal~Daumé III and Aarti Singh, editors, {\em Proceedings of the 37th International Conference on Machine Learning}, volume 119 of {\em Proceedings of Machine Learning Research}, pages 9908--9918. PMLR, 13--18 Jul 2020.

\bibitem{Han2023:MBC}
Shuai Han, Mehdi Dastani, and Shihan Wang.
\newblock Model-based sparse communication in multi-agent reinforcement learning.
\newblock In {\em Proceedings of the 2023 International Conference on Autonomous Agents and Multiagent Systems}, AAMAS '23, page 439–447, Richland, SC, 2023. International Foundation for Autonomous Agents and Multiagent Systems.

\bibitem{Cao2018:Negotiation}
Kris Cao, Angeliki Lazaridou, Marc Lanctot, Joel~Z Leibo, Karl Tuyls, and Stephen Clark.
\newblock Emergent communication through negotiation.
\newblock In {\em International Conference on Learning Representations}, 2018.

\bibitem{Lazaridou2018:Emergence}
Angeliki Lazaridou, Karl~Moritz Hermann, Karl Tuyls, and Stephen Clark.
\newblock Emergence of linguistic communication from referential games with symbolic and pixel input.
\newblock In {\em International Conference on Learning Representations}, 2018.

\bibitem{Jaques2019:SocialInfluence}
Natasha Jaques, Angeliki Lazaridou, Edward Hughes, Caglar Gulcehre, Pedro~A. Ortega, DJ~Strouse, Joel~Z. Leibo, and Nando de~Freitas.
\newblock Social influence as intrinsic motivation for multi-agent deep reinforcement learning.
\newblock In {\em Proceedings of the 36th International Conference on Machine Learning}, pages 3040--3049, 2019.

\bibitem{Kim2018:SchedNet}
Daewoo Kim, Sangwoo Moon, David Hostallero, Wan~Ju Kang, Taeyoung Lee, Kyunghwan Son, and Yung Yi.
\newblock Learning to schedule communication in multi-agent reinforcement learning.
\newblock In {\em International Conference on Learning Representations}, 2019.

\bibitem{Rita2022:GenOverf}
Mathieu Rita, Corentin Tallec, Paul Michel, Jean-Bastien Grill, Olivier Pietquin, Emmanuel Dupoux, and Florian Strub.
\newblock Emergent communication: Generalization and overfitting in lewis games.
\newblock In S.~Koyejo, S.~Mohamed, A.~Agarwal, D.~Belgrave, K.~Cho, and A.~Oh, editors, {\em Advances in Neural Information Processing Systems}, volume~35, pages 1389--1404. Curran Associates, Inc., 2022.

\bibitem{Mordatch2018:GroundedCompo}
Igor Mordatch and Pieter Abbeel.
\newblock Emergence of grounded compositional language in multi-agent populations.
\newblock In Sheila~A. McIlraith and Kilian~Q. Weinberger, editors, {\em Proceedings of the Thirty-Second AAAI Conference on Artificial Intelligence}, pages 1495--1502. {AAAI} Press, 2018.

\bibitem{Rita2022:PopHetero}
Mathieu Rita, Florian Strub, Jean-Bastien Grill, Olivier Pietquin, and Emmanuel Dupoux.
\newblock On the role of population heterogeneity in emergent communication.
\newblock In {\em International Conference on Learning Representations}, 2022.

\bibitem{Chaabouni2019:AntiEfficient}
Rahma Chaabouni, Eugene Kharitonov, Emmanuel Dupoux, and Marco Baroni.
\newblock Anti-efficient encoding in emergent communication.
\newblock In H.~Wallach, H.~Larochelle, A.~Beygelzimer, F.~d\textquotesingle Alch\'{e}-Buc, E.~Fox, and R.~Garnett, editors, {\em Advances in Neural Information Processing Systems}, volume~32. Curran Associates, Inc., 2019.

\bibitem{Lowe2019:Pitfalls}
Ryan Lowe, Jakob Foerster, Y-Lan Boureau, Joelle Pineau, and Yann Dauphin.
\newblock On the pitfalls of measuring emergent communication.
\newblock In {\em Proceedings of the 18th International Conference on Autonomous Agents and MultiAgent Systems}, AAMAS '19, page 693–701, Richland, SC, 2019. International Foundation for Autonomous Agents and Multiagent Systems.

\bibitem{Lazaridou2020_DeepEmergentComm}
Angeliki Lazaridou and Marco Baroni.
\newblock Emergent multi-agent communication in the deep learning era.
\newblock June 2020.

\bibitem{Bouchacourt2018_HowAgentsSee}
Diane Bouchacourt and Marco Baroni.
\newblock How agents see things: On visual representations in an emergent language game.
\newblock In {\em Proceedings of the 2018 Conference on Empirical Methods in Natural Language Processing}, pages 981--985. Association for Computational Linguistics, 2018.

\bibitem{Harnad1990:SymbolGrounding}
Stevan Harnad.
\newblock The symbol grounding problem.
\newblock {\em Physica D: Nonlinear Phenomena}, 42(1):335--346, 1990.

\bibitem{Vogt2002:PhysicalGrounding}
Paul Vogt.
\newblock The physical symbol grounding problem.
\newblock {\em Cognitive Systems Research}, 3(3):429--457, September 2002.

\bibitem{Lee2018:EmergentTranslation}
Jason Lee, Kyunghyun Cho, Jason Weston, and Douwe Kiela.
\newblock Emergent translation in multi-agent communication.
\newblock In {\em International Conference on Learning Representations}, 2018.

\bibitem{Lin2021:GroundMAC}
Toru Lin, Minyoung Huh, Chris Stauffer, Ser-Nam Lim, and Phillip Isola.
\newblock Learning to ground multi-agent communication with autoencoders.
\newblock In {\em Advances in Neural Information Processing Systems}, 2021.

\bibitem{Das2017:CoopVisDial}
Abhishek Das, Satwik Kottur, Jose M.~F. Moura, Stefan Lee, and Dhruv Batra.
\newblock Learning cooperative visual dialog agents with deep reinforcement learning.
\newblock In {\em Proceedings of the IEEE International Conference on Computer Vision (ICCV)}, 2017.

\bibitem{Havrylov2017:EmergenceLang}
Serhii Havrylov and Ivan Titov.
\newblock Emergence of language with multi-agent games: Learning to communicate with sequences of symbols.
\newblock In {\em Proceedings of the 31st International Conference on Neural Information Processing Systems}, NIPS'17, page 2146–2156, Red Hook, NY, USA, 2017. Curran Associates Inc.

\bibitem{Gupta2021:DynamicPop}
Abhinav Gupta, Marc Lanctot, and Angeliki Lazaridou.
\newblock Dynamic population-based meta-learning for multi-agent communication with natural language.
\newblock In A.~Beygelzimer, Y.~Dauphin, P.~Liang, and J.~Wortman Vaughan, editors, {\em Advances in Neural Information Processing Systems}, 2021.

\bibitem{Tucker2021:DiscreteEC}
Mycal Tucker, Huao Li, Siddharth Agrawal, Dana Hughes, Katia~P. Sycara, Michael Lewis, and Julie Shah.
\newblock Emergent discrete communication in semantic spaces.
\newblock In A.~Beygelzimer, Y.~Dauphin, P.~Liang, and J.~Wortman Vaughan, editors, {\em Advances in Neural Information Processing Systems}, 2021.

\bibitem{Lee2019:CounteringLDrift}
Jason Lee, Kyunghyun Cho, and Douwe Kiela.
\newblock Countering language drift via visual grounding.
\newblock In {\em Proceedings of the 2019 Conference on Empirical Methods in Natural Language Processing and the 9th International Joint Conference on Natural Language Processing (EMNLP-IJCNLP)}, pages 4385--4395, Hong Kong, China, 2019. Association for Computational Linguistics.

\bibitem{Lazaridou2020:MACNatLang}
Angeliki Lazaridou, Anna Potapenko, and Olivier Tieleman.
\newblock Multi-agent communication meets natural language: Synergies between functional and structural language learning.
\newblock In {\em Proceedings of the 58th Annual Meeting of the Association for Computational Linguistics}, pages 7663--7674, Online, 2020. Association for Computational Linguistics.

\bibitem{Lowe2020:S2P}
Ryan Lowe, Abhinav Gupta, Jakob Foerster, Douwe Kiela, and Joelle Pineau.
\newblock On the interaction between supervision and self-play in emergent communication.
\newblock In {\em International Conference on Learning Representations}, 2020.

\bibitem{Lazaridou2017}
Angeliki Lazaridou, Alexander Peysakhovich, and Marco Baroni.
\newblock Multi-agent cooperation and the emergence of (natural) language.
\newblock In {\em International Conference on Learning Representations}, 2017.

\bibitem{Karten2023:IMGS-MAC}
Seth Karten, Mycal Tucker, Siva Kailas, and Katia Sycara.
\newblock Towards true lossless sparse communication in multi-agent systems.
\newblock In {\em ICRA 2023}, 2023.

\bibitem{Driess2023:PaLME}
Danny Driess, Fei Xia, Mehdi S.~M. Sajjadi, Corey Lynch, Aakanksha Chowdhery, Brian Ichter, Ayzaan Wahid, Jonathan Tompson, Quan Vuong, Tianhe Yu, Wenlong Huang, Yevgen Chebotar, Pierre Sermanet, Daniel Duckworth, Sergey Levine, Vincent Vanhoucke, Karol Hausman, Marc Toussaint, Klaus Greff, Andy Zeng, Igor Mordatch, and Pete Florence.
\newblock Palm-e: An embodied multimodal language model.
\newblock In PMLR, editor, {\em Proceedings of the 40th International Conference on Machine Learning}, volume 202, 2023.

\bibitem{Carta2023:GLAM}
Thomas Carta, Cl{\'e}ment Romac, Thomas Wolf, Sylvain Lamprier, Olivier Sigaud, and Pierre-Yves Oudeyer.
\newblock {Grounding Large Language Models in Interactive Environments with Online Reinforcement Learning}.
\newblock In {\em Proceedings of Machine Learning Research}, volume 202. {PMLR}, 2023.

\bibitem{Christiano2017:RLHF}
Paul~F Christiano, Jan Leike, Tom Brown, Miljan Martic, Shane Legg, and Dario Amodei.
\newblock Deep reinforcement learning from human preferences.
\newblock In I.~Guyon, U.~Von Luxburg, S.~Bengio, H.~Wallach, R.~Fergus, S.~Vishwanathan, and R.~Garnett, editors, {\em Advances in Neural Information Processing Systems}, volume~30. Curran Associates, Inc., 2017.

\bibitem{Zhu2024:CognitiveLLMs}
Feiyu Zhu and Reid Simmons.
\newblock Bootstrapping cognitive agents with a large language model.
\newblock {\em Proceedings of the AAAI Conference on Artificial Intelligence}, 38(1):655--663, 2024.

\bibitem{Zhang2024:CoELA}
Hongxin Zhang, Weihua Du, Jiaming Shan, Qinhong Zhou, Yilun Du, Joshua~B. Tenenbaum, Tianmin Shu, and Chuang Gan.
\newblock Building cooperative embodied agents modularly with large language models.
\newblock In {\em International Conference on Learning Representations}, 2024.

\bibitem{Wu2023:AutoGen}
Qingyun Wu, Gagan Bansal, Jieyu Zhang, Yiran Wu, Beibin Li, Erkang Zhu, Li~Jiang, Xiaoyun Zhang, Shaokun Zhang, Jiale Liu, Ahmed~Hassan Awadallah, Ryen~W White, Doug Burger, and Chi Wang.
\newblock Autogen: Enabling next-gen llm applications via multi-agent conversation.
\newblock 2023.

\bibitem{Li2023:CAMEL}
Guohao Li, Hasan Hammoud, Hani Itani, Dmitrii Khizbullin, and Bernard Ghanem.
\newblock Camel: Communicative agents for "mind" exploration of large language model society.
\newblock In A.~Oh, T.~Neumann, A.~Globerson, K.~Saenko, M.~Hardt, and S.~Levine, editors, {\em Advances in Neural Information Processing Systems}, volume~36, pages 51991--52008. Curran Associates, Inc., 2023.

\bibitem{Park2023:LLMtown}
Joon~Sung Park, Joseph O'Brien, Carrie~Jun Cai, Meredith~Ringel Morris, Percy Liang, and Michael~S. Bernstein.
\newblock Generative agents: Interactive simulacra of human behavior.
\newblock In {\em Proceedings of the 36th Annual ACM Symposium on User Interface Software and Technology}, UIST '23, New York, NY, USA, 2023. Association for Computing Machinery.

\bibitem{Vezhnevets2023:Concordia}
Alexander~Sasha Vezhnevets, John~P. Agapiou, Avia Aharon, Ron Ziv, Jayd Matyas, Edgar~A. Duéñez-Guzmán, William~A. Cunningham, Simon Osindero, Danny Karmon, and Joel~Z. Leibo.
\newblock Generative agent-based modeling with actions grounded in physical, social, or digital space using concordia.
\newblock 2023.

\bibitem{Perez2024:CulturalEvo}
Jérémy Perez, Corentin Léger, Marcela Ovando-Tellez, Chris Foulon, Joan Dussauld, Pierre-Yves Oudeyer, and Clément Moulin-Frier.
\newblock Cultural evolution in populations of large language models.
\newblock 2024.

\bibitem{Wei2022:ChainOfThought}
Jason Wei, Xuezhi Wang, Dale Schuurmans, Maarten Bosma, brian ichter, Fei Xia, Ed~Chi, Quoc~V Le, and Denny Zhou.
\newblock Chain-of-thought prompting elicits reasoning in large language models.
\newblock In S.~Koyejo, S.~Mohamed, A.~Agarwal, D.~Belgrave, K.~Cho, and A.~Oh, editors, {\em Advances in Neural Information Processing Systems}, volume~35, pages 24824--24837. Curran Associates, Inc., 2022.

\bibitem{Liu2024:HLA}
Jijia Liu, Chao Yu, Jiaxuan Gao, Yuqing Xie, Qingmin Liao, Yi~Wu, and Yu~Wang.
\newblock Llm-powered hierarchical language agent for real-time human-ai coordination.
\newblock In {\em International Conference on Autonomous Agents and Multiagent Systems}, 2024.

\bibitem{Hunt2024:LLMMultiRobot}
William Hunt, Toby Godfrey, and Mohammad~D. Soorati.
\newblock Conversational language models for human-in-the-loop multi-robot coordination.
\newblock In {\em Demonstration at International Conference on Autonomous Agents and Multi-Agent Systems}, 2024.

\bibitem{Acerbi2023:LLMsBias}
Alberto Acerbi and Joseph~M. Stubbersfield.
\newblock Large language models show human-like content biases in transmission chain experiments.
\newblock {\em Proceedings of the National Academy of Sciences}, 120(44), 2023.

\bibitem{Ji2023:Hallucinations}
Ziwei Ji, Nayeon Lee, Rita Frieske, Tiezheng Yu, Dan Su, Yan Xu, Etsuko Ishii, Ye~Jin Bang, Andrea Madotto, and Pascale Fung.
\newblock Survey of hallucination in natural language generation.
\newblock {\em ACM Computing Surveys}, 55(12):1--38, 2023.

\bibitem{fudenberg1998learning}
Drew Fudenberg and David Levine.
\newblock Learning in games.
\newblock {\em European economic review}, 42(3-5):631--639, 1998.

\bibitem{nowak2006evolutionary}
Martin~A Nowak.
\newblock {\em Evolutionary dynamics: exploring the equations of life}.
\newblock Harvard university press, 2006.

\bibitem{sen2014sociophysics}
Parongama Sen and Bikas~K Chakrabarti.
\newblock {\em Sociophysics: an introduction}.
\newblock OUP Oxford, 2014.

\bibitem{galam1982sociophysics}
Serge Galam, Yuval Gefen, and Yonathan Shapir.
\newblock Sociophysics: A new approach of sociological collective behaviour. i. mean-behaviour description of a strike.
\newblock {\em Journal of Mathematical Sociology}, 9(1):1--13, 1982.

\bibitem{tailleur2022active}
Julien Tailleur, Gerhard Gompper, M~Cristina Marchetti, Julia~M Yeomans, and Christophe Salomon.
\newblock {\em Active Matter and Nonequilibrium Statistical Physics: Lecture Notes of the Les Houches Summer School: Volume 112, September 2018}, volume 112.
\newblock Oxford University Press, 2022.

\bibitem{eiben2015introduction}
Agoston~E Eiben and James~E Smith.
\newblock {\em Introduction to evolutionary computing}.
\newblock Springer, 2015.

\bibitem{sutton2018reinforcement}
Richard~S Sutton and Andrew~G. Barto.
\newblock Reinforcement learning: An introduction.
\newblock {\em A Bradford Book}, 2018.

\end{thebibliography}
\bibliographystyle{unsrt}

\end{document}